\documentclass{article}

\usepackage[preprint]{neurips_2022}

\usepackage[utf8]{inputenc} 
\usepackage[T1]{fontenc}    
\usepackage{hyperref}       
\usepackage{url}            
\usepackage{booktabs}       
\usepackage{amsfonts}       
\usepackage{nicefrac}       
\usepackage{microtype}      
\usepackage{xcolor}         

\usepackage{bbm}
\usepackage{amsmath}
\usepackage{multirow}
\usepackage{graphicx}
\usepackage{algorithm}
\usepackage{algpseudocode}
\usepackage{caption}
\usepackage{subcaption}

\newcommand{\R}{\mathbb{R}}
\newcommand{\E}{\mathop{\mathbb{E}}}
\newcommand{\argmin}{\mathop{\text{argmin}}}
\newcommand{\argmax}{\mathop{\text{argmax}}}
\renewcommand{\L}{\mathcal{L}}
\newcommand{\ce}{\mathop{\text{CrossEntropy}}}
\newcommand{\sm}{\mathop{\text{Softmax}}}

\newcommand{\by}{\mathbf{y}}
\newcommand{\bw}{\mathbf{w}}

\newcommand{\bX}{\mathbf{X}}
\newcommand{\bC}{\mathbf{C}}
\newcommand{\bW}{\mathbf{W}}

\usepackage{xcolor}

\newcommand{\white}{\color{white}}

\title{Pick up the PACE: Fast and Simple Domain Adaptation via Ensemble Pseudo-Labeling }

\author{%
    Christopher Liao \\
    Boston University \\
    \texttt{cliao25@bu.edu}
    \And
    Theodoros Tsiligkaridis \\
    MIT Lincoln Laboratory \\
    \texttt{ttsili@ll.mit.edu}
    \And
    Brian Kulis \\
    Boston University \\
    \texttt{bkulis@bu.edu}
}

\begin{document}

\maketitle

\begin{abstract}
    Domain Adaptation (DA) has received widespread attention from deep learning researchers in recent years because of its potential to improve test accuracy with out-of-distribution labeled data. 
    Most state-of-the-art DA algorithms require an extensive amount of hyperparameter tuning and are computationally intensive due to the large batch sizes required. 
    In this work, we propose a fast and simple DA method consisting of three stages: (1) domain alignment by covariance matching, (2) pseudo-labeling, and (3) ensembling. We call this method \textbf{PACE}, for \textbf{P}seudo-labels, \textbf{A}lignment of \textbf{C}ovariances, and \textbf{E}nsembles. PACE is trained on top of \emph{fixed features} extracted from an ensemble of modern pretrained backbones. PACE exceeds previous state-of-the-art by \textbf{5 - 10 \%} on most benchmark adaptation tasks \emph{without training a neural network}. PACE reduces training time and hyperparameter tuning time by 82\% and 97\%, respectively, when compared to state-of-the-art DA methods. 
    Code is released here: \url{https://github.com/Chris210634/PACE-Domain-Adaptation}
\end{abstract}

\section{Introduction}
Deep learning is infamous for requiring a large amount of labeled data to achieve state-of-the-art results, but in many applications, labeled data is expensive to obtain \cite{zhao2020review,zhang2020covid,wang2020covid}. In the last few years, new sub-disciplines of deep learning have emerged to address this over-reliance on labeled data. \emph{Self-supervised learning} learns useful representations of data through handcrafted augmentations instead of labels \cite{jing2020self,jaiswal2020survey}. \emph{Semi-supervised learning} learns from a small amount of labeled data and a large amount of unlabeled data from the same distribution \cite{van2020survey}. \emph{Domain Adaptation (DA)} is similar to semi-supervised learning, but considers the case where the labeled and unlabeled data come from different distributions \cite{zhuang2020comprehensive,wilson2020survey}. We call the labeled data the \emph{source} domain and the unlabeled data the \emph{target} domain. For example, consider the following scenario. A company wants to train a speech recognition system for deployment in a noisy car environment (target domain). A large amount of labeled recordings is available for indoor environments (source domain), but a limited amount of labeled target data is available. DA addresses this problem by minimizing the empirical risk on source data while encouraging features to be domain invariant. We consider both \emph{Semi-supervised Domain Adaptation (SSDA)} \cite{saito2019semi}, where a small amount of labeled target data is available, and \emph{Unsupervised Domain Adaptation (UDA)} \cite{ganin2016domain}, where no labeled target data is available.  

Many modern DA methods train a feature extractor and linear classifier to minimize a weighted sum of three losses: (1) cross entropy loss on labeled source data \cite{ganin2016domain,saito2019semi,acuna2021f,liu2021cycle}, (2) divergence loss between source and target domain features \cite{ganin2016domain,saito2019semi,acuna2021f,zhang2019bridging} and (3) consistency loss between augmented views of unlabeled target data \cite{singh2021clda,li2021cross,zhang2022low}.  
Generally, better performing methods require large batch sizes and a large number of hyperparameters \cite{kim2020attract,li2021cross,liu2021cycle,zhang2022low,singh2021clda}. For example, CDAC \cite{li2021cross}, a state-of-the-art method in SSDA, requires a batch of labeled data, a batch of unlabeled data, and two augmented batches of unlabeled data. On a small feature extractor, such as ResNet34, tuning and running this method is feasible. However, scaling this type of method to large modern feature extractors would require several GPUs working in parallel \cite{liu2021swin,liu2022convnet}.
Computationally efficient training is important for two reasons: (1) there is significant interest in energy-efficient training on resource-constrained edge devices \cite{wang2019e2,he2020group}, and (2) algorithms which achieve state-of-the-art with low computational cost foster democratic research \cite{berthelot2021adamatch}. This raises the question: \emph{is it possible to leverage high quality features from large modern backbones, using only one GPU, and still achieve competitive DA results?} The answer is yes!

\begin{figure}
\centering
\includegraphics[width=1.\linewidth]{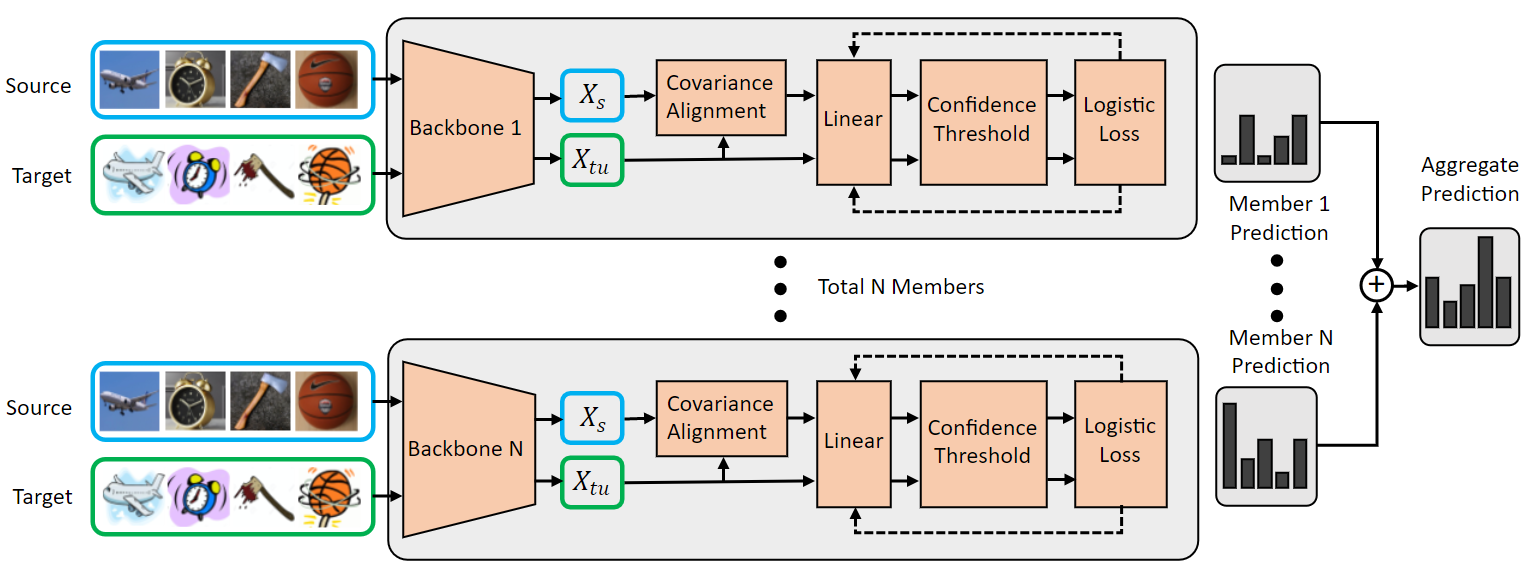}
\caption{Illustration of the PACE framework. A solid arrow indicates a forward pass. A dashed arrow indicates a backward pass. PACE consists of three steps: (1) the source features $\bX_s$ are transformed such that their covariances match the target feature $\bX_{tu}$ covariances, (2) a linear classifier is trained with logistic loss on source data and pseudo-labeled target data, and (3) the predictions from ensemble members are averaged to return an aggregate prediction.}
\label{fig:illustration}
\end{figure}

\noindent \textbf{Present work:} In this paper, we propose an efficient DA method \textbf{PACE} (\textbf{P}seudo-labels, \textbf{A}lignment of \textbf{C}ovariances, and \textbf{E}nsembles), which uses pre-deep learning DA methods on top of \emph{fixed features} extracted from an ensemble of ConvNeXt \cite{liu2022convnet} and Swin \cite{liu2021swin} backbones pre-trained on ImageNet \cite{deng2009imagenet}. Our method consists of three stages (see Figure \ref{fig:illustration}): (1) We align the source and target feature distributions by matching their covariances with CORAL \cite{coral}. (2) We use self-training with confidence thresholding to align the class-conditional feature distributions between source and target domains. (3) We run the first two steps with different pre-trained backbones and average the predictions. We hypothesize that the features extracted by the collection of pre-trained backbones are not perfectly correlated because of differences in pre-training and architecture. Therefore, we should observe a boost in accuracy when combining predictions from the ensemble.

Our contributions are:
\begin{itemize}
    \item We re-visit CORAL \cite{coral}, a simple method for domain alignment that is not widely used by state-of-the-art DA methods. We show that when combined with self-training and ensembling, CORAL offers a boost in target accuracy without adding much complexity.
    \item We propose PACE, which reduces training time and hyperparameter tuning time by 82\% and 97\%, respectively (see Figure \ref{fig:timing-results}) while exceeding state-of-the-art SSDA and UDA methods by 5-10\% on most benchmark tasks (see Tables \ref{tab:UDA-office}, \ref{tab:SSDA-office} and \ref{tab:DomainNet}).
\end{itemize}

\section{Related Work}

\paragraph{Domain Feature Alignment}
Following Ben-David et al.'s DA theory  \cite{ben2010theory}, DANN \cite{ganin2016domain}, MCD \cite{saito2018maximum}, MME \cite{saito2019semi}, MDD \cite{zhang2019bridging}, and f-DAL \cite{acuna2021f} minimize the empirical source risk and some divergence measure between source and target distributions as a surrogate to minimizing target risk. CAN \cite{kang2019contrastive} and CLDA \cite{singh2021clda} use a contrastive approach to align the class-conditional feature distributions. At a high level, these methods train the backbone to output domain-invariant features and improve target accuracy over source-only training. However, the effectiveness of these methods is unclear in the context of large modern backbones, which produce good features without any fine-tuning or adaptation (Table \ref{tab:coral-self-ablation}). A preliminary study \cite{kim2022broad} shows that some state-of-the-art DA methods attain lower target accuracy than source-only training on some benchmark adaptation tasks, due to negative transfer \cite{wang2019characterizing}. It is currently unclear what target accuracy is achievable on benchmarks without any training of the backbone. We fill this gap by presenting results on benchmarks using a fast and simple DA method that does not train the backbone. We hope that these results serve as a helpful baseline for future work.

\paragraph{Pseudo-labeling applied to DA}
Pseudo-labeling \cite{lee2013pseudo} (also called self-training) is a well-established semi-supervised technique that adds artificial supervision to unlabeled data. FixMatch \cite{sohn2020fixmatch} is a popular variant of self-training, which generates high-quality pseudo-labels by confidence thresholding and enforcing consistency between augmented copies of data. Many DA methods use FixMatch as part of their framework \cite{liu2021cycle, li2021cross, zhang2022low, berthelot2021adamatch}, so high quality pseudo-labels is important in the context of DA. Despite the success of these methods, pseudo-labeling is known to suffer from confirmation bias \cite{arazo2020pseudo}. Recent studies try to mitigate this issue with meta-learning \cite{pham2021meta} and debiasing \cite{chen2022debiased}. Our method generates high quality pseudo-labels and can complement pseudo-labeling-based DA methods as the first stage in a DA pipeline. 

\paragraph{Ensembling in DA}
Many application-oriented deep learning studies use an ensemble of multiple deep models to boost accuracy \cite{alshazly2019ensembles, alsafari2020deep, rajaraman2020iteratively}.
Some multi-source DA methods \cite{zhou2021domain, kang2020contrastive} use an ensemble of experts trained on each source domain to obtain more accurate pseudo-labels for target data. Inspired by these studies, we incorporate an ensemble of different backbones as part of our framework and show that it increases target accuracy. We further experiment with forming an ensemble of different augmented views of the dataset. This is inspired by SENTRY \cite{prabhu2021sentry}, which uses a committee of augmented views to guide training. 

\section{Method}
\paragraph{Setup and Notation}
We have a labeled source dataset of size $n_s$, a labeled target dataset of size $n_{tl}$ and an unlabeled target dataset of size $n_{tu}$. The output of the backbone is dimension $d$. We then denote the feature vectors extracted from the backbone as: $\bX_s \in \R^{n_s \times d}, \bX_{tl} \in \R^{n_{tl} \times d}, \bX_{tu} \in \R^{n_{tu} \times d}$. The label vectors are denoted as $\by_s \in \R^{n_s}$ and $\by_{tl} \in \R^{n_{tl}}$ for source and labeled target, respectively. We always optimize over the entire dataset (as opposed to stochastic optimization), so for conciseness, we use the expectation symbol $\E[\cdot]$ to denote the deterministic average. When training linear classifiers, we use logistic regression, which we denote concisely as:
$$\ell(\cdot, y) := \ce(\sm(\cdot), y) $$
\paragraph{Overview}
We form an ensemble from all pretrained ConvNeXt-XL and Swin-L backbones available through the \texttt{timm} \cite{rw2019timm} python library (Table \ref{tab:ensemble-details}) and perform the following steps for \emph{each} ensemble member: (1) Calculate the features for all data; (2) Use CORAL to align the target and source feature distributions; (3) Train a linear classifier on all labeled data; and (4) Fine-tune the linear classifier using self-training to adjust for label shift. Finally, we output the average of ensemble predictions. See Figure \ref{fig:illustration}.
\subsection{CORAL feature alignment}
CORAL \cite{coral} is a popular linear algebraic method for distribution alignment. Its goal is to align labeled and unlabeled feature distributions by transforming the labeled data such that its covariance matrix is the same as the unlabeled data. CORAL consists of two steps. First, we \emph{whiten} the labeled data distribution by subtracting the mean and normalizing covariances. Note that for this step, we concatenate the source data and labeled target data together:
\begin{equation}
\begin{split}
    \bX_{\text{labeled}} &= \text{concatenate}(\bX_s, \bX_{tl}) \\
    \bX_{\text{labeled}} &= \bX_{\text{labeled}} - \E[\bX_{\text{labeled}}] \: , \: \bX_{tu} = \bX_{tu} - \E[\bX_{tu}] \\
    \bX_{\text{labeled}} &= \bX_{\text{labeled}} \cdot \bC_{\text{labeled}}^{-1/2} \: \text{ where } \bC_{\text{labeled}} = \text{Cov}(\bX_{\text{labeled}}) + \lambda I
\end{split}
\label{eq:coral1}
\end{equation}
where $\lambda$ is a small numerical constant to ensure the covariance matrix is invertible. $\bX_{\text{labeled}}$ now looks like samples from a standard normal distribution with zero-mean and unit variance. Next, we \emph{re-color} the source data with the target covariance matrix:
\begin{equation}
    \bX_{\text{labeled}} = \bX_{\text{labeled}} \cdot \bC_{tu}^{1/2} \: \text{ where } \bC_{tu} = \text{Cov}(\bX_{tu}) + \lambda I
\label{eq:coral2}
\end{equation}
The labeled and unlabeled distributions are now aligned in the sense that it is impossible to train a binary linear classifier which distinguishes the two distributions. We would hope that training a linear classifier on the recolored labeled data would yield similar accuracy between labeled and unlabeled data. However, this is not the case because of \emph{label shift}. Namely, the source and target distributions, when conditioned on the labels, are not the same. In the next subsection, we use \emph{self-training} to train a linear classifier which accounts for label shift. 

\subsection{Self Training}
Following prior DA literature \cite{saito2019semi,singh2021clda,li2021cross}, we L2-normalize the features, then train a zero-bias linear classifier on the labeled data with logistic loss. The resulting classifier is biased towards the source distribution. To adjust for this bias, we pseudo-label the target features using the source classifier and calculate confidences. We then fine-tune the classifier with the confident source and target samples, using simple confidence thresholding. The resulting classifier achieves higher target accuracy at the expense of lower source accuracy. Importantly, note that we discard source samples with low confidence, because they are unlikely to be informative of target data. Concretely, let $\bW$ be the weight matrix of the linear classifier being trained; $\bW \in \R^{K \times d}$, where $K$ is the number of classes. We repeat the following for $T$ iterations ($t \in 1:T$):
\begin{equation}
\begin{split}
    \text{Pseudo-labeling:} \:\:\:\: & \hat{\by}_{tu} = \argmax \bX_{tu} \bW^\top \\
    \text{Confidence Thresholding:} \:\:\:\: & \text{mask}_s = \mathbbm{1}[\bX_s \bW^\top > \tau_t^s] \: , \: \text{mask}_{tu} = \mathbbm{1}[\bX_{tu} \bW^\top > \tau_t^{tu}] \\
    \text{Train classifier:}\:\:\:\: & \bW = \argmin_{\bW} \alpha_t \L_s + \beta_t \L_{tl} + \gamma_t \L_{tu} \\
    \L_s &:= \E[\ell(\bX_s \bW^\top, \by_s) \cdot \text{mask}_s] \\
    \L_{tl} &:= \E[\ell(\bX_{tl} \bW^\top, \by_{tl})] \\
    \L_{tu} &:= \E[\ell(\bX_{tu} \bW^\top, \hat{\by}_{tu}) \cdot \text{mask}_{tu} ]
\end{split}
\label{eq:self-train}
\end{equation}
where $\tau_t^s$ and $\tau_t^{tu}$ are confidence thresholds for source and unlabeled target data at iteration $t$. $\alpha_t$, $\beta_t$, and $\gamma_t$ are hyperparameters. We solve the $\argmin$ over $\bW$ in Equation \ref{eq:self-train} using Gradient Descent (GD). The pseudo-labeling and classifier fine-tuning steps are repeated for a fixed number of iterations $T$. This iterative approach gradually adapts the linear classifier to the target distribution.

\subsection{Ensembling}
It is well known that aggregating the predictions of weak classifiers results in a strong classifier, under the assumption that the predictions of the weak classifiers are not completely correlated (Chapter 14 of \cite{bishop2006pattern}). We hypothesize that the classifiers trained on each backbone do not make completely correlated predictions. Hence, averaging the predictions from an ensemble of backbones should outperform the best member backbone. We demonstrate experimentally in Table \ref{tab:ablation-augment} that the average prediction of the ensemble is almost always better than the best ensemble member.

\begin{algorithm}
\caption{PACE}\label{alg:cap}
\textbf{Input: } Labeled target data $\bX_{tl}$, Unlabeled target data $\bX_{tu}$, Labeled source data $\bX_s$\\
\textbf{Output: } Predictions for unlabeled target data $\hat{\by}_{tu}$
\\
\\
\textbf{Step 1:} CORAL feature alignment. (Do for each ensemble member)
\\
Modify $\bX_s$ , $\bX_{tl}$ and $\bX_{tu}$ according to Equations \ref{eq:coral1} and \ref{eq:coral2}.
\\
\\
\textbf{Step 2:} Train labeled classifier. (Do for each ensemble member)
\\
L2-Normalize $\bX_s$, $\bX_{tl}$, and $\bX_{tu}$.
\\
Initialize weight matrix $\bW$, then using GD with learning rate $\eta_0$, solve:
\\
$$\bW = \argmin_{\bW} \alpha_0 \L_{s} + \beta_0 \L_{tl},\: \text{where } \L_s := \E[\ell(\bX_s \bW^\top, \by_s)],\: \L_{tl} := \E[\ell(\bX_{tl} \bW^\top, \by_{tl})]$$
\\
\textbf{Step 3:} Self Training. (Do for each ensemble member)
\\
\textbf{For} $t$ in $T$ iterations:
\\
{\white inde} Use GD with learning rate $\eta_t$ to train $\bW$ following Equation \ref{eq:self-train}. 
\\
\\
\textbf{Step 4:} Ensemble by averaging. 
\\
With some abuse of notation, we denote $\bW_i$ as the weights of the $i$-th ensemble member.
$$\hat{\by}_{tu} = \argmax\sum_{i} \sm (\bX_{tu} \bW_i^\top ) $$
\end{algorithm}

\begin{table}
\caption{ Details on ensemble members}
\label{tab:ensemble-details}
\centering
{ 
\begin{tabular}{ c | c c c c c c  }
 \toprule
 \multirow{2}{*}{Model} & Input  & Output  & \multirow{2}{*}{Pre-train} & \multirow{2}{*}{Fine-tune} & \multicolumn{2}{c}{Throughput (img/sec)} \\
  & Size & Size &  & & P100 & V100 \\
 \hline
ConvNeXt-XL\cite{liu2022convnet} & $384^2$ & 2048 & ImageNet 22K & ImageNet 1K & 13.9 & 27.3 \\
ConvNeXt-XL\cite{liu2022convnet} & $224^2$ & 2048 & ImageNet 22K & ImageNet 1K & 38.2 & 67.6 \\
ConvNeXt-XL\cite{liu2022convnet} & $224^2$ & 2048 & ImageNet 22K & None & 38.2 & 67.8 \\
Swin-L\cite{liu2021swin} & $224^2$ & 1536 & ImageNet 22K & ImageNet 1K & 66.4 & 103.6 \\
Swin-L\cite{liu2021swin} & $224^2$ & 1536 & ImageNet 22K & None & 64.6 & 91.7 \\
Swin-L\cite{liu2021swin} & $384^2$ & 1536 & ImageNet 22K & ImageNet 1K & 22.1 & 37.1 \\
Swin-L\cite{liu2021swin} & $384^2$ & 1536 & ImageNet 22K & None & 22.0 & 37.6 \\
 \bottomrule
\end{tabular}
}
\end{table}

\section{Experiments}
\paragraph{Datasets}
We validate our method on two common domain adaptation benchmarks: \textbf{Office-Home} \cite{venkateswara2017deep} and \textbf{DomainNet} \cite{peng2019moment}. Office-Home contains 15,588 images belonging to 64 classes from 4 domains: Art, Product, Real, and Clipart. Following SSDA literature, we use a subset of DomainNet \cite{saito2019semi}, which contains 145,145 images belonging to 126 classes from 4 domains: Art, Painting, Real, and Clipart. For Office-Home, we present UDA results in Table \ref{tab:UDA-office} and 3-shot SSDA results in Table \ref{tab:SSDA-office} on all 12 source-target combinations. For DomainNet, we present 1-shot and 3-shot SSDA results in Table \ref{tab:DomainNet} on 7 source-target combinations used in prior SSDA work. 1-shot means that only 1 labeled target sample per class is used; 3-shot means that 3 labeled target samples per class are used.

\paragraph{Baselines}
We compare our UDA performance on Office-Home with baselines in Table \ref{tab:UDA-office}. All baselines are cited inline. All baselines use ResNet-50 backbone with ImageNet-1K pretraining unless otherwise noted. We also include comparisons to recent transformer-based methods, which achieve higher target accuracy than traditional ResNet-based approaches. Some papers report results on transformer networks of varying size; we report the best result reported by the authors. We compare our SSDA performance with baselines in Tables \ref{tab:SSDA-office} and \ref{tab:DomainNet}. All SSDA baselines use ResNet-34 pretrained on ImageNet-1K.

\paragraph{Hyperparameters}
We use the same set of hyperparameters across all datasets and tasks. Each ensemble member is trained with the same hyperparameters using  Algorithm 1. We use GD on the entire dataset with momentum=0.9 with Nestorov correction. We do not regularize classifier weights. We tune the learning rates $\eta_0 ... \eta_T$ and number of GD iterations such that the loss converges to a reasonable value. $\eta_0 = 40$, $\eta_{1:T} = 80$. Initial training on labeled data (Step 2) runs for 400 GD iterations, all self training steps (Step 3) run for 200 GD iterations. The remaining hyperparameters are tuned coarsely on the target validation accuracy of the SSDA 3-shot DomainNet Real to Clipart task. The initial guess for hyperparameter values is guided by the intuition that as training progresses, the classifier should rely less on source labels and more on target pseudo-labels. The final values used for all results are: $T=30$, $\alpha_0 = 0.4$, $\beta_0 = 0.2$, $\alpha_{1:T} = 0.1$, $\beta_{1:T} = 0.05$, $\gamma_{1:T} = 0.9$, $\tau^s_{1:T} = 0.8$, $\tau^{tu}_{1:10} = 0.9$, $\tau^{tu}_{11:20} = 0.8$, $\tau^{tu}_{21:30} = 0.7$. See Appendix for results using different hyperparameters.

We include code and results for each ensemble member in the supplementary material.

\paragraph{Performance Metrics}
All results are top-1 target accuracy. Following prior work, UDA target accuracy is reported over the entire target dataset (there is no training/validation/test split). In the SSDA setting, target data is split into labeled, validation and unlabeled sets. Target accuracy is reported over the unlabeled set. The validation set contains 3 target samples per class and is only used for tuning hyperparameters.

\setlength\tabcolsep{1.5 pt}
\begin{table}
\caption{UDA results on Office-Home. All baselines use ResNet-50\cite{he2016deep} as backbone unless otherwise noted. WinTR uses DeiT-S\cite{touvron2021training}. CDTrans uses DeiT-B\cite{touvron2021training}. TVT and SSRT use ViT-B\cite{dosovitskiy2020image}. BCAT uses Swin-B\cite{liu2021swin}.
Ours (small) refers to the ensemble of 7 backbones on unaugmented data. Ours (large) refers to the ensemble of 28 (7 backbones $\times$ 4 augmented views). 
The domains are abbreviated as A:Art, C:Clipart, P:Product, R:Real. }
\centering
{ \small
\begin{tabular}{ l | c c c c c c c c c c c c | c }
 \toprule
 Method & A - C & A - P & A - R & C - A & C - P & C - R & P - A & P - C & P - R & R - A & R - C & R - P & Mean\\
 \hline
Source-Only & 34.9  & 50.0  & 58.0  & 37.4  & 41.9  & 46.2  & 38.5  & 31.2  & 60.4  & 53.9  & 41.2  & 59.9  & 46.1  \\
DANN (JMLR '16) \cite{ganin2016domain} & 45.6  & 59.3  & 70.1  & 47.0  & 58.5  & 60.9  & 46.1  & 43.7  & 68.5  & 63.2  & 51.8  & 76.8  & 57.6  \\
JAN (ICML '17) \cite{long2017deep} & 45.9  & 61.2  & 68.9  & 50.4  & 59.7  & 61.0  & 45.8  & 43.4  & 70.3  & 63.9  & 52.4  & 76.8  & 58.3  \\
CDAN (Neurips '18) \cite{long2018conditional} & 50.7  & 70.6  & 76.0  & 57.6  & 70.0  & 70.0  & 57.4  & 50.9  & 77.3  & 70.9  & 56.7  & 81.6  & 65.8  \\
ALDA (AAAI '20) \cite{chen2020adversarial} & 53.7  & 70.1  & 76.4  & 60.2  & 72.6  & 71.5  & 56.8  & 51.9  & 77.1  & 70.2  & 56.3  & 82.1  & 66.6  \\
SAFN (ICCV '19) \cite{xu2019larger} & 52.0  & 71.7  & 76.3  & 64.2  & 69.9  & 71.9  & 63.7  & 51.4  & 77.1  & 70.9  & 57.1  & 81.5  & 67.3  \\
SymNet (CVPR '19) \cite{zhang2019domain} & 47.7  & 72.9  & 78.5  & 64.2  & 71.3  & 74.2  & 64.2  & 48.8  & 79.5  & 74.5  & 52.6  & 82.7  & 67.6  \\
TADA (AAAI '19) \cite{wang2019transferable} & 53.1  & 72.3  & 77.2  & 59.1  & 71.2  & 72.1  & 59.7  & 53.1  & 78.4  & 72.4  & 60.0  & 82.9  & 67.6  \\
MDD (ICML '19) \cite{zhang2019bridging} & 54.9  & 73.7  & 77.8  & 60.0  & 71.4  & 71.8  & 61.2  & 53.6  & 78.1  & 72.5  & 60.2  & 82.3  & 68.1  \\
BNM (CVPR '20) \cite{cui2020towards} & 56.2  & 73.7  & 79.0  & 63.1  & 73.6  & 74.0  & 62.4  & 54.8  & 80.7  & 72.4  & 58.9  & 83.5  & 69.4  \\
MDD+IA (ICML '20) \cite{jiang2020implicit} & 56.2  & 77.9  & 79.2  & 64.4  & 73.1  & 74.4  & 64.2  & 54.2  & 79.9  & 71.2  & 58.1  & 83.1  & 69.5  \\
f-DAL (ICML '21) \cite{acuna2021f} & 56.7  & 77.0  & 81.1  & 63.1  & 72.2  & 75.9  & 64.5  & 54.4  & 81.0  & 72.3  & 58.4  & 83.7  & 70.0  \\
CADA-P (CVPR '19) \cite{kurmi2019attending} & 56.9  & 76.4  & 80.7  & 61.3  & 75.2  & 75.2  & 63.2  & 54.5  & 80.7  & 73.9  & 61.5  & 84.1  & 70.2  \\
GSDA (CVPR '20) \cite{hu2020unsupervised} & 61.3  & 76.1  & 79.4  & 65.4  & 73.3  & 74.3  & 65.0  & 53.2  & 80.0  & 72.2  & 60.6  & 83.1  & 70.3  \\
GVB (CVPR '20) \cite{cui2020gradually} & 57.0  & 74.7  & 79.8  & 64.6  & 74.1  & 74.6  & 65.2  & 55.1  & 81.0  & 74.6  & 59.7  & 84.3  & 70.4  \\
DCAN (AAAI '20) \cite{li2020domain} & 54.5  & 75.7  & 81.2  & 67.4  & 74.0  & 76.3  & 67.4  & 52.7  & 80.6  & 74.1  & 59.1  & 83.5  & 70.5  \\
TCM (ICCV '21) \cite{yue2021transporting} & 58.6  & 74.4  & 79.6  & 64.5  & 74.0  & 75.1  & 64.6  & 56.2  & 80.9  & 74.6  & 60.7  & 84.7  & 70.7  \\
HDAN (Neurips '20) \cite{cui2020heuristic} & 56.8  & 75.2  & 79.8  & 65.1  & 73.9  & 75.2  & 66.3  & 56.7  & 81.8  & 75.4  & 59.7  & 84.7  & 70.9  \\
MetaAlign (CVPR '21) \cite{wei2021metaalign} & 59.3  & 76.0  & 80.2  & 65.7  & 74.7  & 75.1  & 65.7  & 56.5  & 81.6  & 74.1  & 61.1  & 85.2  & 71.3  \\
ToAlign (Neurips '21) \cite{wei2021toalign} & 57.9  & 76.9  & 80.8  & 66.7  & 75.6  & 77.0  & 67.8  & 57.0  & 82.5  & 75.1  & 60.0  & 84.9  & 72.0  \\
SENTRY (ICCV '21) \cite{prabhu2021sentry} & 61.8  & 77.4  & 80.1  & 66.3  & 71.6  & 74.7  & 66.8  & 63.0  & 80.9  & 74.0  & 66.3  & 84.1  & 72.2  \\
FixBi (CVPR '21) \cite{na2021fixbi} & 58.1  & 77.3  & 80.4  & 67.7  & 79.5  & 78.1  & 65.8  & 57.9  & 81.7  & 76.4  & 62.9  & 86.7  & 72.7  \\
CST (Neurips '21) \cite{liu2021cycle} & 59.0  & 79.6  & 83.4  & 68.4  & 77.1  & 76.7  & 68.9  & 56.4  & 83.0  & 75.3  & 62.2  & 85.1  & 73.0  \\
SCDA (ICCV '21) \cite{li2021semantic} & 60.7  & 76.4  & 82.8  & 69.8  & 77.5  & 78.4  & 68.9  & 59.0  & 82.7  & 74.9  & 61.8  & 84.5  & 73.1  \\
ATDOC (CVPR '21) \cite{liang2021domain} & 60.2  & 77.8  & 82.2  & 68.5  & 78.6  & 77.9  & 68.4  & 58.4  & 83.1  & 74.8  & 61.5  & 87.2  & 73.2  \\
PCL (arxiv '21) \cite{li2021semanticPCL} & 60.8  & 79.8  & 81.6  & 70.1  & 78.9  & 78.9  & 69.9  & 60.7  & 83.3  & 77.1  & 66.4  & 85.9  & 74.5  \\
MixLRCo (arxiv '22) \cite{zhang2022low} & 64.4  & 81.1  & 81.6  & 68.5  & 78.9  & 78.8  & 69.1  & 59.9  & 87.0  & 77.3  & 67.7  & 86.7  & 75.1  \\
WinTR (arxiv '21) \cite{ma2021exploiting} & 65.3  & 84.1  & 85.0  & 76.8  & 84.5  & 84.4  & 73.4  & 60.0  & 85.7  & 77.2  & 63.1  & 86.8  & 77.2  \\
CDTrans (ICLR '22) \cite{xu2021cdtrans} & 68.8  & 85.0  & 86.9  & 81.5  & 87.1  & 87.3  & 79.6  & 63.3  & 88.2  & 82.0  & 66.0  & 90.6  & 80.5  \\
TVT (arxiv '21) \cite{yang2021tvt} & 74.9  & 86.8  & 89.5  & 82.8  & 88.0  & 88.3  & 79.8  & 71.9  & 90.1  & 85.5  & 74.6  & 90.6  & 83.6  \\
SSRT (CVPR '22) \cite{sun2022safe} & 75.2  & 89.0  & 91.1  & 85.1  & 88.3  & 90.0  & 85.0  & 74.2  & 91.3  & 85.7  & 78.6  & 91.8  & 85.4  \\
BCAT (arxiv '22) \cite{wang2022domain} & 75.3  & 90.0  & 92.9  & 88.6  & 90.3  & 92.7  & 87.4  & 73.7  & 92.5  & 86.7  & 75.4  & 93.5  & 86.6  \\
 \hline
Ours (small) & 82.5  & \textbf{94.2}  & 94.2  & 90.3  & \textbf{93.7}  & 94.2  & 89.3  & 82.0  & 94.5  & 89.8  & 85.0  & 95.1  & 90.4  \\
Ours (large) & \textbf{84.0}  & 94.0  & \textbf{94.3}  & \textbf{90.7}  & 93.6  & \textbf{94.4}  & \textbf{89.6}  &\textbf{83.0}  & \textbf{94.6}  & \textbf{90.3}  & \textbf{85.7}  & \textbf{95.3}  & \textbf{90.8}  \\
\bottomrule
\end{tabular}
}
\label{tab:UDA-office}
\end{table}

\setlength\tabcolsep{1.4 pt}
\begin{table}
\caption{SSDA results on Office-Home. 3-shot. All baselines use ResNet-34. S+T refers to training on only labeled data (no adaptation).}
\centering
{ \small
\begin{tabular}{ l | c c c c c c c c c c c c | c }
 \toprule
 Method & R - C & R - P & R - A & P - R & P - C & P - A & A - P & A - C & A - R & C - R & C - A & C - P & Mean\\
 \hline
S+T  & 55.7  & 80.8  & 67.8  & 73.1  & 53.8  & 63.5  & 73.1  & 54.0  & 74.2  & 68.3  & 57.6  & 72.3  & 66.2  \\
DANN (JMLR '16) \cite{ganin2016domain} & 57.3  & 75.5  & 65.2  & 69.2  & 51.8  & 56.6  & 68.3  & 54.7  & 73.8  & 67.1  & 55.1  & 67.5  & 63.5  \\
ENT (ICCV '19) \cite{saito2019semi} & 62.6  & 85.7  & 70.2  & 79.9  & 60.5  & 63.9  & 79.5  & 61.3  & 79.1  & 76.4  & 64.7  & 79.1  & 71.9  \\
MME (ICCV '19) \cite{saito2019semi} & 64.6  & 85.5  & 71.3  & 80.1  & 64.6  & 65.5  & 79.0  & 63.6  & 79.7  & 76.6  & 67.2  & 79.3  & 73.1  \\
Meta-MME (ECCV '20) \cite{li2020online} & 65.2  & -  & -  & -  & 64.5  & 66.7  & -  & 63.3  & -  & -  & 67.5  & -  & -  \\
APE (ECCV '20) \cite{kim2020attract} & 66.4  & 86.2  & 73.4  & 82.0  & 65.2  & 66.1  & 81.1  & 63.9  & 80.2  & 76.8  & 66.6  & 79.9  & 74.0  \\
CDAC (CVPR '21) \cite{li2021cross} & 67.8  & 85.6  & 72.2  & 81.9  & 67.0  & 67.5  & 80.3  & 65.9  & 80.6  & 80.2  & 67.4  & 81.4  & 74.2  \\
CLDA (Neurips '21) \cite{singh2021clda} & 66.0  & 87.6  & 76.7  & 82.2  & 63.9  & 72.4  & 81.4  & 63.4  & 81.3  & 80.3  & 70.5  & 80.9  & 75.5  \\
SSDAS (arxiv '21) \cite{huang2021semi} & 69.1  & 86.9  & 76.2  & 83.4  & 66.8  & 67.5  & 83.5  & 63.8  & 82.3  & 77.9  & 67.0  & 81.1  & 75.5  \\
DECOTA (ICCV '21) \cite{yang2021deep} & 70.4  & 87.7  & 74.0  & 82.1  & 68.0  & 69.9  & 81.8  & 64.0  & 80.5  & 79.0  & 68.0  & 83.2  & 75.7  \\
MCL (arxiv '22) \cite{yan2022multi} & 70.1  & 88.1  & 75.3  & 83.0  & 68.0  & 69.9  & 83.9  & 67.5  & 82.4  & 81.6  & 71.4  & 84.3  & 77.1  \\
PCL (arxiv '21) \cite{li2021semanticPCL} & 69.1  & 89.5  & 76.9  & 83.8  & 68.0  & 74.7  & 85.5  & 67.6  & 82.3  & 82.7  & 73.4  & 83.4  & 78.1  \\
MixLRCo (arxiv '22) \cite{zhang2022low} & 73.1  & 89.6  & 77.0  & 84.2  & 71.3  & 73.5  & 84.5  & 70.2  & 83.2  & 83.1  & 72.0  & 86.0  & 79.0  \\
\hline
Ours (small) & 86.1	& 95.6	& 90.5	& \textbf{95.4}	& 84.5	& 90.2	& \textbf{95.5}	& 85.3	& \textbf{95.2}	& \textbf{94.9}	& 90.4	& \textbf{95.3}	& 91.6 \\
Ours (large) & \textbf{87.0}	& \textbf{95.7}	& \textbf{90.8}	& 95.1	& \textbf{85.0}	& \textbf{90.7}	& 95.3	& \textbf{86.3}	& 94.9	& \textbf{94.9}	& \textbf{91.2}	& \textbf{95.3}	& \textbf{91.9} \\
 \bottomrule
\end{tabular}
}
\label{tab:SSDA-office}
\end{table}
\setlength\tabcolsep{6 pt}

\setlength\tabcolsep{2.5 pt}
\begin{table}
\caption{SSDA Results on DomainNet. All baselines use ResNet-34. The domains are abbreviated as R:Real, C:Clipart, P:Painting, S:Sketch. 1-sh means 1-shot. 3-sh means 3-shot.}
\centering
{ \small
\begin{tabular}{ l | c c | c c | c c | c c | c c | c c | c c | c c  }
 \toprule
\multirow{2}{*}{Method} & \multicolumn{2}{c}{R $\rightarrow$ C} & \multicolumn{2}{c}{R $\rightarrow$ P} & \multicolumn{2}{c}{P $\rightarrow$ C} & \multicolumn{2}{c}{C $\rightarrow$ S} & \multicolumn{2}{c}{S $\rightarrow$ P} & \multicolumn{2}{c}{R $\rightarrow$ S} & \multicolumn{2}{c}{P $\rightarrow$ R}& \multicolumn{2}{c}{Mean}\\
 & 1-sh & 3-sh & 1-sh & 3-sh & 1-sh & 3-sh & 1-sh & 3-sh & 1-sh & 3-sh & 1-sh & 3-sh & 1-sh & 3-sh & 1-sh & 3-sh\\
 \hline
S+T & 55.6  & 60.0  & 60.6  & 62.2  & 56.8  & 59.4  & 50.8  & 55.0  & 56.0  & 59.5  & 46.3  & 50.1  & 71.8  & 73.9  & 56.9  & 60.0  \\
DANN \cite{ganin2016domain} & 58.2  & 59.8  & 61.4  & 62.8  & 56.3  & 59.6  & 52.8  & 55.4  & 57.4  & 59.9  & 52.2  & 54.9  & 70.3  & 72.2  & 58.4  & 60.7  \\
ENT \cite{saito2019semi} & 65.2  & 71.0  & 65.9  & 69.2  & 65.4  & 71.1  & 54.6  & 60.0  & 59.7  & 62.1  & 52.1  & 61.1  & 75.0  & 78.6  & 62.6  & 67.6  \\
MME \cite{saito2019semi} & 70.0  & 72.2  & 67.7  & 69.7  & 69.0  & 71.7  & 56.3  & 61.8  & 64.8  & 66.8  & 61.0  & 61.9  & 76.1  & 78.5  & 66.4  & 68.9  \\
BiAT \cite{jiang2020bidirectional} & 73.0  & 74.9  & 68.0  & 68.8  & 71.6  & 74.6  & 57.9  & 61.5  & 63.9  & 67.5  & 58.5  & 62.1  & 77.0  & 78.6  & 67.1  & 69.7  \\
Meta-MME \cite{li2020online} & -  & 73.5  & -  & 70.3  & -  & 72.8  & -  & 62.8  & -  & 68.0  & -  & 63.8  & -  & 79.2  & -  & 70.1  \\
UODA \cite{qin2021contradictory} & 72.7  & 75.4  & 70.3  & 71.5  & 69.8  & 73.2  & 60.5  & 64.1  & 66.4  & 69.4  & 62.7  & 64.2  & 77.3  & 80.8  & 68.5  & 71.2  \\
Li et al. \cite{li2021graph} & 72.8  & 75.5  & 71.2  & 72.5  & 69.2  & 73.9  & 60.3  & 63.6  & 65.4  & 68.2  & 63.7  & 66.4  & 77.1  & 80.0  & 68.5  & 71.4  \\
Con2DA \cite{perez20222} & 71.3  & 74.2  & 71.8  & 72.1  & 71.1  & 75.0  & 60.0  & 65.7  & 63.5  & 67.1  & 65.2  & 67.1  & 75.7  & 78.6  & 68.4  & 71.4  \\
APE \cite{kim2020attract} & 70.4  & 76.6  & 70.8  & 72.1  & 72.9  & 76.7  & 56.7  & 63.1  & 64.5  & 66.1  & 63.0  & 67.8  & 76.6  & 79.4  & 67.6  & 71.7  \\
S3D \cite{yoon2022semi} & 73.3  & 75.9  & 68.9  & 72.1  & 73.4  & 75.1  & 60.8  & 64.4  & 68.2  & 70.0  & 65.1  & 66.7  & 79.5  & 80.3  & 69.9  & 72.1  \\
ATDOC \cite{liang2021domain} & 74.9  & 76.9  & 71.3  & 72.5  & 72.8  & 74.2  & 65.6  & 66.7  & 68.7  & 70.8  & 65.2  & 64.6  & 81.2  & 81.2  & 71.4  & 72.4  \\
DFA \cite{zhang2021dfa} & 71.8  & 76.7  & 72.7  & 73.9  & 69.8  & 75.4  & 60.8  & 65.5  & 68.0  & 70.5  & 62.3  & 67.5  & 76.8  & 80.3  & 68.9  & 72.8  \\
ToAlilgn \cite{wei2021toalign} & 73.0  & 75.7  & 72.0  & 72.9  & 71.7  & 75.6  & 63.0  & 66.3  & 69.3  & 71.1  & 64.6  & 66.4  & 80.8  & 83.0  & 70.6  & 73.0  \\
STar \cite{singh2021improving} & 74.1  & 77.1  & 71.3  & 73.2  & 71.0  & 75.8  & 63.5  & 67.8  & 66.1  & 69.2  & 64.1  & 67.9  & 80.0  & 81.2  & 70.0  & 73.2  \\
PAC \cite{mishra2021surprisingly} & 74.9  & 78.6  & 73.0  & 74.3  & 72.6  & 76.0  & 65.8  & 69.6  & 67.9  & 69.4  & 68.7  & 70.2  & 76.7  & 79.3  & 71.4  & 73.9  \\
CLDA \cite{singh2021clda} & 76.1  & 77.7  & 75.1  & 75.7  & 71.0  & 76.4  & 63.7  & 69.7  & 70.2  & 73.7  & 67.1  & 71.1  & 80.1  & 82.9  & 71.9  & 75.3  \\
DECOTA \cite{yang2021deep} & -  & 80.4  & -  & 75.2  & -  & 78.7  & -  & 68.6  & -  & 72.7  & -  & 71.9  & -  & 81.5  & -  & 75.6  \\
CDAC \cite{li2021cross} & 77.4  & 79.6  & 74.2  & 75.1  & 75.5  & 79.3  & 67.6  & 69.9  & 71.0  & 73.4  & 69.2  & 72.5  & 80.4  & 81.9  & 73.6  & 76.0  \\
ECACL \cite{li2021ecacl} & 75.3  & 79.0  & 74.1  & 77.3  & 75.3  & 79.4  & 65.0  & 70.6  & 72.1  & 74.6  & 68.1  & 71.6  & 79.7  & 82.4  & 72.8  & 76.4  \\
MCL \cite{yan2022multi} & 77.4  & 79.4  & 74.6  & 76.3  & 75.5  & 78.8  & 66.4  & 70.9  & 74.0  & 74.7  & 70.7  & 72.3  & 82.0  & 83.3  & 74.4  & 76.5  \\
PCL \cite{li2021semanticPCL} & 78.1  & 80.5  & 75.2  & 78.1  & 77.2  & 80.3  & 68.8  & 74.1  & 74.5  & 76.5  & 70.1  & 73.5  & 81.9  & 84.1  & 75.1  & 78.2  \\
MixLRCo \cite{zhang2022low} & 78.7  & 81.9  & 76.9  & 77.5  & 78.3  & 81.2  & 68.5  & 74.4  & 74.2  & 75.3  & 72.8  & \textbf{75.6}  & 81.1  & 83.5  & 75.8  & 78.5  \\
\hline
Ours (small) & 82.3  & 84.0  & 84.2  & 84.7  & 82.2  & 84.2  & 73.6  & 75.2  & 84.6  & 85.1  & 72.8  & 75.0  & \textbf{91.7}  & \textbf{92.5}  & 81.6  & 83.0  \\
Ours (large) & \textbf{82.4}  & \textbf{84.2}  & \textbf{84.5}  & \textbf{84.9}  & \textbf{82.6}  & \textbf{84.5}  & \textbf{74.6}  & \textbf{76.0}  & \textbf{84.8}  & \textbf{85.3}  & \textbf{74.0}  & 75.4  & \textbf{91.7}  & \textbf{92.5}  & \textbf{82.1}  &\textbf{83.3}  \\
\bottomrule
\end{tabular}
}
\label{tab:DomainNet}
\end{table}
\setlength\tabcolsep{6 pt}

\subsection{Domain Adaptation Results}

We form an ensemble with the 7 pretrained backbones listed in Table \ref{tab:ensemble-details}. We report results from the ensemble of 7 backbones as ``Ours (small)'' in the tables. We further experiment with augmenting the dataset to obtain multiple different views. We use 4 views of the data: (1) the original data (2) perspective-preserving resize, by padding extra space with zeros, (3) RandAugment \cite{cubuk2020randaugment} and (4) gray-scale. Using the 7 $\times$ 4 combinations of backbone and augmentation, we form a larger ensemble of 28 members. These results are reported as ``Ours (large)''.

From Tables \ref{tab:UDA-office}, \ref{tab:SSDA-office} and \ref{tab:DomainNet}, both our large and small ensembles clearly beat the state-of-the-art by a healthy margin on all benchmark tasks except one. We emphasize that we achieve this gain with a simpler, faster, and easier-to-tune method. Our method's performance is heavily influenced by what the \emph{target} images look like. In domains where the images look similar to ImageNet (Product and Real), our ensemble achieves very high target accuracy. Our ensemble's target accuracy is lower in more "cartoonish" domains such as Clipart, Art and Sketch.

The larger 7 $\times$ 4 ensemble improves target accuracy marginally but consistently. An average of $0.3 \%$ improvement is achieved. The larger ensemble improves target accuracy significantly when the target domain is Art or Clipart or Sketch. This is likely because the augmentations cause the classifier to rely less heavily on features of real images learned from ImageNet pretraining. In practice, if the best choice for augmentation is known, it may be best to fix the augmentation and run the small ensemble to save time.

\begin{figure}
\centering
\begin{subfigure}{.32\textwidth}
  \centering
  \includegraphics[width=1.\linewidth]{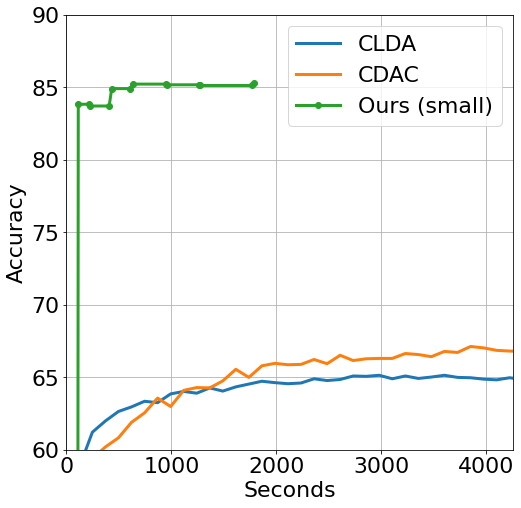}
\end{subfigure}
\begin{subfigure}{.32\textwidth}
  \centering
  \includegraphics[width=1.\linewidth]{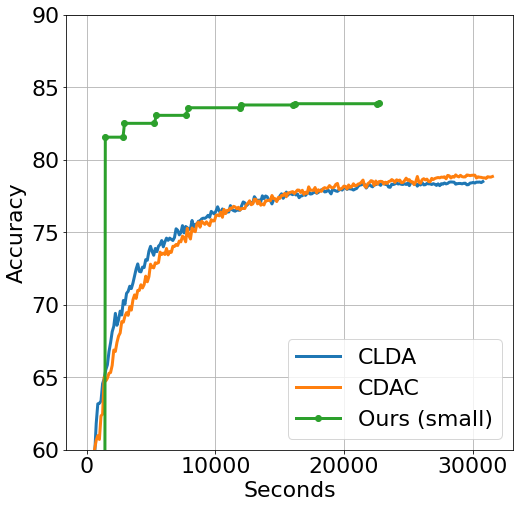}
\end{subfigure}
\begin{subfigure}{.32\textwidth}
  \centering
  \includegraphics[width=1.\linewidth]{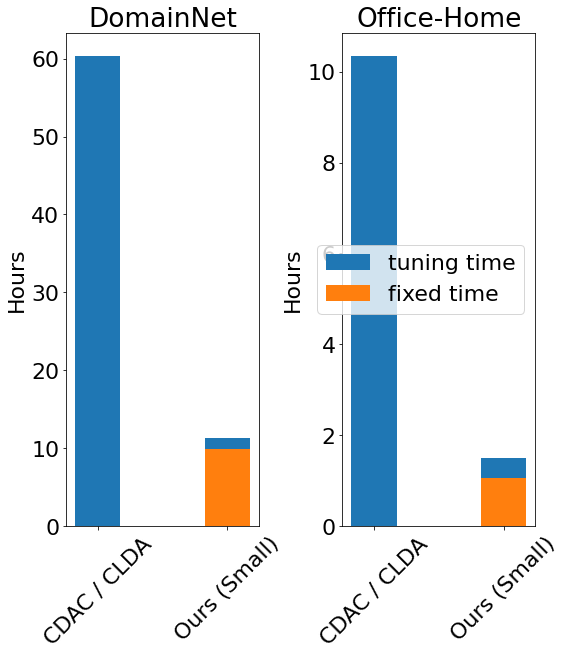}
\end{subfigure}
\caption{Timing results. (Left) Accuracy over time for 3-shot Art to Clipart on Office-Home. (Center Left) Accuracy over time for 3-shot Real to Clipart on DomainNet. (Center Right) Total time needed to complete all 7 3-shot DomainNet adaptation tasks. (Right) Total time needed to complete all 12 3-shot Office-Home adaptation tasks. Time for our method is divided into inference stage in orange, which only needs to be run once, and training stage in blue, which needs to be tuned. Times are reported on a single P100 GPU.}
\label{fig:timing-results}
\end{figure}

\paragraph{Timing Results}
One of the advantages of our framework is that it achieves state-of-the-art accuracy faster than previous state-of-the-art. We demonstrate this in Figure \ref{fig:timing-results} as a comparison between our small ensemble and CDAC and CLDA on DomainNet and Office-Home. For the purpose of this comparison, we order the ensemble members from low to high computational cost and run Algorithm 1 in this order. After the completion of each ensemble member, we check the target accuracy of the average prediction and report this in the plot. We observe that our small ensemble achieves higher target accuracy regardless of the computational budget. On the right side of Figure \ref{fig:timing-results} we compare the total time to run all 3-shot DomainNet and Office-Home SSDA results between CDAC / CLDA and our small ensemble. Note that CLDA takes longer than CDAC because of larger batch sizes, but for the purpose of this experiment, we modify CLDA to use the same amount of time as CDAC without sacrificing accuracy. CDAC and CLDA take  60.3 P100 GPU-hours to complete all 7 3-shot DomainNet tasks compared to 11.3 P100 GPU-hours for our small ensemble. Similarly, CDAC and CLDA take 10.3 P100 GPU-hours to complete all 12 3-shot Office-Home tasks compared to 1.5 hours for our small ensemble. Furthermore, only the training of linear classifiers needs to be tuned for our method. Hence, only 0.4 hours and 1.4 hours of our method needs tuning, for Office-Home and DomainNet, respectively. Therefore, tuning our method takes 2-4\% of the time it takes to tune a method like CDAC or CLDA. This is an enormous practical advantage.

\vspace{-0.5em}
\subsection{Ablation Experiments}
We conduct an extensive set of ablation experiments to justify each component of our method.

\vspace{-0.5em}
\paragraph{CORAL and self-training}
Table \ref{tab:coral-self-ablation} demonstrates the contribution of CORAL and self-training to our method (Steps 1 and 3 in Algorithm 1). Note that CORAL by itself does not consistently improve target accuracy over the naive source classifier. Self-training by itself offers a 1.2\% gain on the Office-Home benchmark and a 3.9\% gain on the DomainNet benchmark. On both benchmarks, self-training with CORAL achieves higher target accuracy than self-training by itself.

\vspace{-0.5em}
\paragraph{Ensembling}
Table \ref{tab:ablation-augment} shows that the average prediction of the ensemble is almost always better than the \emph{best} target accuracy of any ensemble member. This result holds regardless of whether all augmentations are used (``full ensemble'' in Table \ref{tab:ablation-augment}) or if only one of the augmented datasets is used. This is a powerful result which justifies the additional compute time required for the ensemble. Note that the full 4 $\times$ 7 ensemble achieves better or equal numerical results on all but 3 Office-Home tasks compared to the ensemble with only the non-augmented view. The optimal augmentation varies across tasks, but the full ensemble is able to closely match the best out of 4 target accuracies.

\setlength\tabcolsep{2 pt}
\begin{table}
\caption{CORAL and self-training ablations. No Adaption means training a linear classifier on labeled data only. All experiments use the small ensemble with no augmentations.}
\centering
{ \small
\begin{tabular}{ l | c c c c c c c c c c c c | c }
 \toprule
Office-Home (3-shot) & R - C & R - P & R - A & P - R & P - C & P - A & A - P & A - C & A - R & C - R & C - A & C - P & Mean\\
 \hline
No Adaptation & 84.2 & 94.9 & 90.0 & 94.1 & 82.7 & 88.9 & 94.3 & 82.8 & 93.7 & 93.7 & 89.9 & 94.3 & 90.3 \\
CORAL Only & 84.4 & 94.7 & 89.2 & 94.0 & 82.2 & 88.4 & 93.9 & 82.8 & 93.7 & 92.9 & 89.2 & 93.9 & 89.9 \\
Self-Training Only & 85.7 & \textbf{95.8} & \textbf{91.1} & 95.4 & \textbf{85.0} & 89.9 & 94.8 & \textbf{85.8} & 94.9 & 94.8 & \textbf{90.4} & 94.8 & 91.5 \\
CORAL + Self-Training & \textbf{86.1} & 95.6 & 90.5 & \textbf{95.4} & 84.5 & \textbf{90.2} & \textbf{95.5} & 85.3 & \textbf{95.2} & \textbf{94.9} & \textbf{90.4} & \textbf{95.3} & \textbf{91.6} \\
 \bottomrule
\end{tabular}
\setlength\tabcolsep{6.29 pt}
\begin{tabular}{ l | c c c c c c c | c }
 \toprule
 DomainNet (3-shot)  & R $\rightarrow$ C & R $\rightarrow$ P & P $\rightarrow$ C & C $\rightarrow$ S & S $\rightarrow$ P & R $\rightarrow$ S & P $\rightarrow$ R & Mean\\
 \hline
No Adaptation & 79.0 & 82.8 & 80.5 & 73.7 & 83.4 & 70.4 & 91.9 & 78.3 \\
CORAL Only & 78.7 & 81.7 & 78.7 & 71.8 & 81.2 & 71.5 & 90.5 & 79.2 \\
Self-Training Only & 83.6 & 84.2 & \textbf{84.6} & 73.7 & 84.4 & 72.2 & \textbf{92.8} & 82.2 \\
CORAL + Self-Training & \textbf{84.0} & \textbf{84.7} & 84.2 & \textbf{75.2} & \textbf{85.1} & \textbf{75.0} & 92.5 & \textbf{83.0} \\
\bottomrule
\end{tabular}
}
\label{tab:coral-self-ablation}
\end{table}
\setlength\tabcolsep{6 pt}

\setlength\tabcolsep{1.3 pt}
\begin{table}
\caption{Comparison of augmentations. ``Averaged'' refers to an ensemble formed by the 7 backbones on the same augmented view. }
\centering
{ \small
\begin{tabular}{ l | c c c c c c c c c c c c | c }
\toprule
\multicolumn{14}{c}{Office-Home (3-shot)} \\
 \midrule
 Augmentation & R - C & R - P & R - A & P - R & P - C & P - A & A - P & A - C & A - R & C - R & C - A & C - P & Mean \\
 \hline
Grayscale (Best Member) & 82.8 & 95.2 & 89.9 & 94.2 & 81.2 & 88.8 & 94.6 & 82.8 & 93.5 & 93.2 & 88.8 & 93.9 & 89.5 \\
Grayscale (Averaged) & 84.0 & 94.5 & 90.0 & 94.5 & 82.4 & 89.0 & 93.8 & 84.0 & 94.1 & 93.9 & 89.6 & 94.2 & 90.3 \\
 \hline
Perspective (Best Member) & 86.1 & 95.7 & 90.0 & 94.7 & 84.2 & 89.7 & 95.1 & 85.8 & 94.3 & 94.3 & 90.4 & 94.5 & 91.0 \\
Perspective (Averaged) & \textbf{87.5} & 95.6 & 90.7 & 95.0 & \textbf{85.1} & 90.3 & 95.1 & \textbf{86.8} & 94.7 & 94.7 & 91.0 & 95.1 & 91.8 \\
 \hline
RandAugment (Best Member) & 84.1 & 95.0 & 90.3 & 94.7 & 82.6 & 89.2 & 94.9 & 83.4 & 94.1 & 93.9 & 89.5 & 94.5 & 90.2 \\
RandAugment (Averaged) & 86.0 & 95.5 & \textbf{90.9} & 95.1 & 84.5 & \textbf{90.8} & 95.1 & 85.3 & 95.0 & \textbf{95.0} & 90.1 & 95.1 & 91.5 \\
 \hline
None (Best Member) & 85.0 & \textbf{95.8} & 90.6 & 94.9 & 83.2 & 89.4 & 95.4 & 84.1 & 94.6 & 94.1 & 89.6 & 94.7 & 90.6 \\
None (Averaged) & 86.1 & 95.6 & 90.5 & \textbf{95.4} & 84.5 & 90.2 & \textbf{95.5} & 85.3 & \textbf{95.2} & 94.9 & 90.4 & \textbf{95.3} & 91.6 \\
\hline
Full Ensemble & 87.0 & 95.7 & 90.8 & 95.1 & 85.0 & 90.7 & 95.3 & 86.3 & 94.9 & 94.9 & \textbf{91.2} & \textbf{95.3} & \textbf{91.9} \\
 \bottomrule
 
\end{tabular}
\setlength\tabcolsep{5.2pt}
\begin{tabular}{ l | c c c c c c c | c }
\toprule
\multicolumn{9}{c}{DomainNet (3-shot)} \\
 \midrule
 Augmentation & R $\rightarrow$ C & R $\rightarrow$ P & P $\rightarrow$ C & C $\rightarrow$ S & S $\rightarrow$ P & R $\rightarrow$ S & P $\rightarrow$ R & Mean\\
 \hline
Grayscale (Best Member) & 79.6 & 82.2 & 79.5 & 75.8 & 83.0 & 74.4 & 90.3 & 80.7 \\
Grayscale (Averaged) & 81.3 & 82.6 & 81.2 & 76.0 & 83.2 & 75.0 & 90.6 & 81.4 \\
\hline
Perspective (Best Member) & 82.2 & 84.0 & 83.1 & 74.2 & 84.4 & 74.1 & 92.1 & 81.7 \\
Perspective (Averaged) & 83.7 & 84.3 & \textbf{84.5} & 74.7 & 84.8 & 74.1 & 92.3 & 82.6 \\
\hline
RandAugment (Best Member) & 82.5 & 83.8 & 82.8 & 74.1 & 83.9 & 73.3 & 92.0 & 81.3 \\
RandAugment (Averaged) & 84.1 & 84.6 & 84.4 & 75.2 & 85.0 & 74.9 & \textbf{92.5} & 82.9 \\
\hline
None (Best Member) & 83.0 & 84.0 & 83.0 & 74.5 & 84.2 & 73.9 & 92.1 & 81.7 \\
None (Averaged) & 84.0 & 84.7 & 84.2 & 75.2 & 85.1 & 75.0 & \textbf{92.5} & 83.0 \\
\hline
Full Ensemble & \textbf{84.2} & \textbf{84.9} & \textbf{84.5} & \textbf{76.0} & \textbf{85.3} & \textbf{75.4} & \textbf{92.5} & \textbf{83.3} \\
 \bottomrule
\end{tabular}
}
\label{tab:ablation-augment} 
\end{table}
\setlength\tabcolsep{6 pt}

\vspace{-0.5em}
\section{Limitations}
\vspace{-0.5em}
Our work is limited in the following ways: (1) Using a network pre-trained on ImageNet may not work well for certain domain-specific datasets such as synthetic aperture radar, or medical imaging. Furthermore, it would be interesting to perform additional studies on less-curated, in-the-wild datasets using pre-trained models. (2) This work does not consider inference time cost constraints. Inference using an ensemble of large backbones could be impractical in some situations. It would be interesting to investigate using the predictions from our method as pseudo-labels to train a smaller network (e.g. ResNet-34). 

\vspace{-0.5em}
\section{Conclusion}
\vspace{-0.5em}
We propose PACE, a fast and simple Domain Adaptation method, which achieves superior target accuracy on DA benchmarks, while drastically reducing training and tuning time. PACE is built on top of CORAL, self-training, and ensembling, but is to the best of our knowledge the first to investigate a combination of the three methods in a unified framework for both SSDA and UDA.

\vspace{-0.5em}
\section*{Broader Impact}
\vspace{-0.5em}
The proposed methodology for domain adaptation lowers the computational cost and barrier for entry for ML practitioners while achieving superior DA performance, and thus may be used in novel applications (e.g. edge computing) where computational cost is a concern. DA is important for applications where labels are expensive or difficult to obtain, for example, medical imaging, bioinformatics, self-driving vehicles, languages with few native speakers, and recommendation systems. Also, the ability to work without target labels has the potential to enhance privacy.

Though our work has shown very successful domain transfer results, we note that social biases associated with feature representations require further study to gain an improved understanding of model behavior. 

\section*{Acknowledgments}
DISTRIBUTION STATEMENT A. Approved for public release. Distribution is unlimited.

This material is based upon work supported by the Under Secretary of Defense for Research and Engineering under Air Force Contract No. FA8702-15-D-0001. Any opinions, findings, conclusions or recommendations expressed in this material are those of the author(s) and do not necessarily reflect the views of the Under Secretary of Defense for Research and Engineering.

\bibliographystyle{unsrt}
\bibliography{main}

\newpage
\appendix

\section{Code}
Our code is released on GitHub (\url{https://github.com/Chris210634/PACE-Domain-Adaptation}). There is no license. Instructions are included with the code. We ran the experiments on a computing cluster with P100 and V100 GPUs. Our job was assigned either a single P100 or V100 GPU depending on availability.

\section{Assets Used}
We use DomainNet \cite{peng2019moment} and Office-Home \cite{venkateswara2017deep} datasets. Some of our code is adapted from \cite{saito2019semi} (\url{https://github.com/VisionLearningGroup/SSDA_MME}). These assets are not licensed.

\section{Additional Experimental Results}

We include 1-shot SSDA results on Office-Home in Table \ref{tab:SSDA-office-1shot}. Very few baselines are listed because most prior work do not publish 1-shot results on Office-Home with the ResNet backbone.

\paragraph{Multi-run Results} We include multi-run results for 3-shot Office-Home, 3-shot DomainNet, and zero-shot Office-Home (UDA) in Table \ref{tab:SSDA-office-multirun}, Table \ref{tab:SSDA-domainnet-multirun} and Table \ref{tab:UDA-office-multirun}, respectively. All multi-run results are for the small ensemble on the un-augmented dataset. The standard deviation is with respect to three trials. In the SSDA setting, each trial uses a different random labeled/unlabeled split of the target dataset. in the UDA setting, there is no labeled target data, so the standard deviation is only with respect to randomness in hardware (recall that Algorithm 1 is deterministic).

\subsection{Comparison to Domain Adaptation \emph{with} backbone training}
We have constrained our method to training a linear classifier on top of fixed features to reduce the computation required. However, how does this compare to DA with backbone training? Kim et al. \cite{kim2022broad} experiment with standard UDA methods on three Office-Home adaptation tasks using ConvNeXt-XL and Swin-L backbones pretrained on ImageNet-22k. For ease of comparison, we copy relevant results from Table 4 in \cite{kim2022broad} into Table \ref{tab:compare-backbone-training}. In this comparison, our method outperforms the \emph{best} combination of backbone-adaptation method \emph{without} training the backbone. This is a promising but preliminary result. We note that the parameters used in \cite{kim2022broad} may not be optimal, and the exact pretrained backbone used may not be the same as in our study. We also emphasize that we focus on the resource-constrained case and leave any investigation of backbone training to future work. 

\subsection{Comparison to Bootstrap Aggregation (Bagging)}
Bagging is a well-known traditional ensembling method (see Section 14.2 of \cite{bishop2006pattern}). In bagging, we make multiple ``bootstrapped'' copies of the dataset by sampling with replacement, which guarantees that that the copies are not identical. We then run Algorithm 1 on each bootstrapped dataset and return the average prediction. We compare an ensemble formed by 7 bootstrapped datasets to our small ensemble formed by 7 backbones in Table \ref{tab:bootstrap}. We use the ConvNeXt-XL backbone with $384^2$ input resolution and ImageNet-1K fine-tuning for the bagging experiment. From Table \ref{tab:bootstrap}, observe that an ensemble of bootstrapped datasets performs better than no ensemble at all, but our ensemble of different backbones is clearly better than bagging.

\setlength\tabcolsep{2.2 pt}
\begin{table}
\caption{SSDA results on Office-Home (1-shot)}
\centering
{
\begin{tabular}{ l | c c c c c c c c c c c c | c }
 \toprule
 Method & R - C & R - P & R - A & P - R & P - C & P - A & A - P & A - C & A - R & C - R & C - A & C - P & Mean\\
 \hline
 CLDA \cite{singh2021clda} & 60.2 & 83.2 & 72.6 & 81.0 & 55.9 & 66.2 & 76.1 & 56.3 & 79.3 & 76.3 & 66.3 &  73.9 & 70.6 \\
 DECOTA \cite{yang2021deep} & 47.2 &  80.3 & 64.6 & 75.5 & 47.2 & 56.6 & 71.1 & 42.5 & 73.1 & 71.0 & 57.8 & 72.9 & 63.3 \\
 MCL \cite{yan2022multi} & 67.0 & 85.5 & 73.8 & 81.3 & 61.1 & 68.0 & 79.5 & 64.4 & 81.2 & 78.4 & 68.5 & 79.3 & 74.0 \\
 \hline
Ours (small) & 85.7	& \textbf{95.2}	& 90.0	& 94.5	& 82.6	& 90.2	& 95.1	& 84.6	& 94.3	& 94.1	& 90.6	& \textbf{94.9}	& 91.0 \\
Ours (large)  & \textbf{86.2}	& \textbf{95.2}	& \textbf{90.4}	& \textbf{94.7}	& \textbf{82.9}	& \textbf{90.4}	& \textbf{95.2}	& \textbf{85.4}	& \textbf{94.4}	& \textbf{94.3}	& \textbf{91.1}	& 94.7	& \textbf{91.2} \\
 \bottomrule
\end{tabular}
}
\label{tab:SSDA-office-1shot}
\end{table}
\setlength\tabcolsep{6 pt}

\setlength\tabcolsep{2.2 pt}
\begin{table}
\caption{Multi-run results for Office-Home (3-shot). Standard deviation based on 3 trials.}
\centering
{ \tiny
\begin{tabular}{ l c c c c c c c c c c c c c }
 \toprule
 Method & R - C & R - P & R - A & P - R & P - C & P - A & A - P & A - C & A - R & C - R & C - A & C - P & Mean\\
 \midrule
Ours (small)  & 85.9$\pm$0.1	& 95.7$\pm$0.1	& 90.7$\pm$0.5	& 94.9$\pm$0.3	& 84.8$\pm$0.9	& 90.4$\pm$0.5	& 95.5$\pm$0.1	& 85.3$\pm$0.6	& 94.9$\pm$0.3	& 94.8$\pm$0.1	& 90.7$\pm$0.5	& 95.6$\pm$0.3	& 91.6$\pm$0.3 \\
 \bottomrule
\end{tabular}
}
\label{tab:SSDA-office-multirun}
\end{table}
\setlength\tabcolsep{6 pt}

\begin{table}
\caption{Multi-run results for DomainNet (3-shot). Standard deviation based on 3 trials.}
\centering
\setlength\tabcolsep{4.5 pt}
{ \small
\begin{tabular}{ l c c c c c c c c }
\toprule
 Method & R $\rightarrow$ C & R $\rightarrow$ P & P $\rightarrow$ C & C $\rightarrow$ S & S $\rightarrow$ P & R $\rightarrow$ S & P $\rightarrow$ R & Mean\\
 \midrule
Ours (small) & 84.1$\pm$0.2	& 84.6$\pm$0.2	& 83.7$\pm$0.6	& 75.0$\pm$0.2	& 85.1$\pm$0.4	& 74.4$\pm$0.1	& 92.2$\pm$0.1	& 82.7$\pm$0.1 \\
 \bottomrule
\end{tabular}
}
\label{tab:SSDA-domainnet-multirun}
\end{table}
\setlength\tabcolsep{6 pt}

\setlength\tabcolsep{2.2 pt}
\begin{table}
\caption{Multi-run results for Office-Home (UDA). Standard deviation based on 3 trials.}
\centering
{ \tiny
\begin{tabular}{ l c c c c c c c c c c c c c }
 \toprule
 Method & R - C & R - P & R - A & P - R & P - C & P - A & A - P & A - C & A - R & C - R & C - A & C - P & Mean\\
 \midrule
Ours (small) & 84.4$\pm$0.2	& 95.1$\pm$0.1	& 90.0$\pm$0.1	& 94.8$\pm$0.1	& 82.7$\pm$0.1	& 88.9$\pm$0.1	& 94.1$\pm$0.0	& 82.5$\pm$0.0	& 94.4$\pm$0.0	& 94.3$\pm$0.0	& 90.2$\pm$0.1	& 93.5$\pm$0.1	& 90.4$\pm$0.1 \\
 \bottomrule
\end{tabular}
}
\label{tab:UDA-office-multirun}
\end{table}
\setlength\tabcolsep{6 pt}

\begin{table}
\caption{ Comparison to DA with backbone training \cite{kim2022broad}.}
\centering
{
\begin{tabular}{ c c | c | c c c | c  }
 \toprule
& Backbone & Adaptation & R-A & R-C & R-P & Mean \\
\hline
\multirow{12}{*}{\rotatebox[origin=c]{90}{Numbers from Table 4 in \cite{kim2022broad}}} & Swin-L & Source-only & 74.3 & 83.4 & 90.9 & 82.8 \\
& ConvNeXt-XL & Source-only & 74.0 & \textbf{85.1} & 91.4 & 83.5  \\
& Swin-L & DANN \cite{ganin2016domain} & 87.3 & 79.5 & 93.0 & 86.6 \\
& ConvNeXt-XL & DANN \cite{ganin2016domain} & 87.2 & 79.8 & 93.1 & 86.7  \\
& Swin-L & CDAN \cite{long2018conditional} & 90.1 & 81.9 & 93.1 & 88.4  \\
& ConvNeXt-XL & CDAN \cite{long2018conditional} & \textbf{90.2} & 84.6 & 93.8 & \textbf{89.5}  \\
& Swin-L & AFN \cite{xu2019larger} & 87.4 & 77.6 & 92.0 & 85.7 \\
& ConvNeXt-XL & AFN \cite{xu2019larger} & 86.0 & 77.7 & 92.8 & 85.5  \\
& Swin-L & MDD \cite{zhang2019bridging} & 87.8 & 78.0 & 93.6 & 86.5 \\
& ConvNeXt-XL & MDD \cite{zhang2019bridging} & 88.0 & 77.6 & 93.6 & 86.4  \\
& Swin-L & MCC \cite{jin2020minimum} & 89.6 & 81.1 & 94.1 & 88.3  \\
& ConvNeXt-XL & MCC \cite{jin2020minimum} & 89.5 & 82.9 & \textbf{94.4} & 88.9  \\
 \hline
& \multicolumn{2}{c}{Max} & 90.2 & 85.1 & 94.4 & 89.5 \\ 
 \hline
& \multicolumn{2}{c}{Ours (small) (Table \ref{tab:UDA-office})} & 89.8 & 85.0 & 95.1 & 90.0 \\
& \multicolumn{2}{c}{Ours (large) (Table \ref{tab:UDA-office})} & \textbf{90.3} & \textbf{85.7} & \textbf{95.3} & \textbf{90.4} \\
\bottomrule
\end{tabular}
}
\label{tab:compare-backbone-training}
\end{table}

\begin{table}
\caption{Bagging. Comparison of an ensemble formed by 7 bootstrapped datasets to our small ensemble. Only the mean target accuracy for each benchmark is included for brevity. No augmentations are used. 3-shot.}
\centering
{
\begin{tabular}{ l  c c  }
 \toprule
 Mean target accuracy & Office-Home & DomainNet \\
 \midrule
No ensemble (ConvNeXt-XL-384) & 90.6 & 81.7 \\
Our small ensemble (7 members) & \textbf{91.6} & \textbf{83.0} \\
Bagging (7 bootstrapped datasets) & 90.9 & 81.9\\
 \bottomrule
\end{tabular}
}
\label{tab:bootstrap}
\end{table}

\subsection{Projecting Away Domain Direction (PADD)}
We introduce PADD as a plug-in alternative to CORAL for domain alignment. PADD finds a linear domain-invariant transformation of the features by iteratively projecting away the ``domain direction''. Concretely, for each iteration $t \in 1:T$, we train a binary classifier to discriminate between labeled samples and unlabeled samples. Let $\bw \in \R^d$, we solve:
\begin{equation}
    \bw_t = \argmin_{\bw} \frac{1}{2} \E[\ell_{\text{BCE}}(\sigma(\bX_{\text{labeled}}\bw), \mathbf{0})] + \frac{1}{2} \E[\ell_{\text{BCE}}(\sigma(\bX_{tu}\bw), \mathbf{1})] + \lambda_{\text{L1}}\|\bw\|_1
\label{eq:PADD1}
\end{equation}
Similar to the main paper, we optimize Equation \ref{eq:PADD1} over the entire dataset using GD. Here, we run GD for 200 iterations with learning rate 4.0, momentum 0.9, no L2 regularization, L1 regularization $\lambda_{\text{L1}} = 0.0002$, and with Nestorov correction. $\ell_{\text{BCE}}$ denotes the binary cross-entropy loss and $\sigma$ denotes the sigmoid function. We then remove the component of the feature vector that points in the direction $\bw_t$:
\begin{equation}
\begin{split}
    \bX_{\text{labeled}} &= \bX_{\text{labeled}} - \text{proj}_{\bw_t} \bX_{\text{labeled}} \\ 
    \bX_{tu} &= \bX_{tu} - \text{proj}_{\bw_t} \bX_{tu}
\end{split}
\label{eq:PADD2}
\end{equation}
where $\text{proj}_{\bw_t} \bX$ denotes the projection of the rows of $\bX$ onto the vector $\bw_t$. This process is repeated for $T=30$ iterations. In the end, it is no longer possible to discriminate between labeled and unlabeled data using a linear classifier, i.e. $\E[\sigma(\bX_{\text{labeled}}\bw_T)] = \E[\sigma(\bX_{tu}\bw_T)] = 0.5$.
One advantage of PADD is that it is more amenable to mini-batching compared to CORAL; one disadvantage is that PADD may take slightly longer to run. Table \ref{tab:PADD} shows that PADD attains similar target accuracies as CORAL in our framework.

\subsection{Combining Backbone Features without Ensembling}
We introduce an alternative method for combining features from different backbones that does not involve ensembling. We first concatenate the features to form one high-dimensional feature vector. We then run the CORAL and self-training steps on the first $k$ principle components of the high-dimensional feature vector. All hyperparameters are the same as the main paper. Results for varying $k$ are presented in Table \ref{tab:PCA-CORAL}. We make the following observations: (1) There is no clear advantage to setting $k>2048$. (2) This PCA-CORAL-self-training procedure is clearly better than CORAL and self-training on the best backbone (``best ensemble member'' in the table). (3) We still obtain better numerical results with the ensembling procedure outlined in the main paper. PCA-CORAL-self-training is faster than training an ensemble.

\subsection{Comparison of Different Ensembling Strategies}
In the main paper, we consider combining the ensemble predictions by simple averaging. There are more creative ways of combining the predictions, such as majority voting, majority voting weighted by confidence, and majority voting weighted by confidence and validation accuracy. We list results for each of these strategies in Table \ref{tab:Different-ensembling-strategies}. None of these more creative ensembling strategies are better than simple averaging.

\setlength\tabcolsep{2 pt}
\begin{table}
\caption{Comparison of PADD vs. CORAL with and without self-training. No Adaption means training a linear classifier on labeled data only. All experiments use the small ensemble with no augmentations.}
\centering
{ \small
\begin{tabular}{ l | c c c c c c c c c c c c | c }
 \toprule
Office-Home (3-shot) & R - C & R - P & R - A & P - R & P - C & P - A & A - P & A - C & A - R & C - R & C - A & C - P & Mean\\
 \hline
No Adaptation & 84.2 & 94.9 & 90.0 & 94.1 & 82.7 & 88.9 & 94.3 & 82.8 & 93.7 & 93.7 & 89.9 & 94.3 & 90.3 \\
CORAL Only & 84.4 & 94.7 & 89.2 & 94.0 & 82.2 & 88.4 & 93.9 & 82.8 & 93.7 & 92.9 & 89.2 & 93.9 & 89.9 \\
PADD Only & 85.1 & 95.2 & 89.8 & 94.2 & 83.5 & 89.2 & 94.4 & 83.4 & 93.8 & 93.9 & 89.7 & 94.1 & 90.5 \\
Self-Training Only & 85.7 & 95.8 & \textbf{91.1} & 95.4 & 85.0 & 89.9 & 94.8 & \textbf{85.8} & 94.9 & 94.8 & 90.4 & 94.8 & 91.5 \\
CORAL + Self-Training & 86.1 & 95.6 & 90.5 & \textbf{95.4} & 84.5 & \textbf{90.2} & \textbf{95.5} & 85.3 & \textbf{95.2} & \textbf{94.9} & 90.4 & 95.3 & 91.6 \\
PADD +  Self-Training & \textbf{86.5} & \textbf{95.9} & \textbf{91.1} & 95.1 & \textbf{85.3} & 90.1 & 95.1 & 85.5 & 95.0 & \textbf{94.9} & \textbf{90.6} & \textbf{95.5} & \textbf{91.7} \\
\bottomrule
\end{tabular}
\setlength\tabcolsep{6.29 pt}
\begin{tabular}{ l | c c c c c c c | c }
 \toprule
 DomainNet (3-shot)  & R $\rightarrow$ C & R $\rightarrow$ P & P $\rightarrow$ C & C $\rightarrow$ S & S $\rightarrow$ P & R $\rightarrow$ S & P $\rightarrow$ R & Mean\\
 \hline
No Adaptation & 79.0 & 82.8 & 80.5 & 73.7 & 83.4 & 70.4 & 91.9 & 78.3 \\
CORAL Only & 78.7 & 81.7 & 78.7 & 71.8 & 81.2 & 71.5 & 90.5 & 79.2 \\
PADD Only & 79.4 & 82.7 & 80.2 & 72.7 & 82.5 & 71.5 & 91.6 & 80.1 \\
Self-Training Only & 83.6 & 84.2 & 84.6 & 73.7 & 84.4 & 72.2 & \textbf{92.8} & 82.2 \\
CORAL + Self-Training & 84.0 & \textbf{84.7} & 84.2 & \textbf{75.2} & \textbf{85.1} & \textbf{75.0} & 92.5 & \textbf{83.0} \\
PADD + Self-Training & \textbf{84.1} & 84.5 & \textbf{84.8} & 74.1 & 84.2 & 72.4 & 92.7 & 82.4 \\
\bottomrule
\end{tabular}
}
\label{tab:PADD}
\end{table}
\setlength\tabcolsep{6 pt}

\setlength\tabcolsep{2 pt}
\begin{table}
\caption{Comparison of PCA-CORAL-self-training with varying $k$ to ensembling. }
\centering
{ \small
\begin{tabular}{ l | c c c c c c c c c c c c | c }
\toprule
\multicolumn{14}{c}{Office-Home (3-shot) No Augmentation} \\
 \midrule
 Method & R - C & R - P & R - A & P - R & P - C & P - A & A - P & A - C & A - R & C - R & C - A & C - P & Mean \\
 \hline
Best ensemble member & 85.0 & \textbf{95.8} & \textbf{90.6} & 94.9 & 83.2 & 89.4 & 95.4 & 84.1 & 94.6 & 94.1 & 89.6 & 94.7 & 90.6 \\
PCA, $k=1024$ & 85.9 & 95.5 & 90.3 & 94.7 & 84.3 & 90.5 & \textbf{95.5} & 84.9 & 94.8 & 94.1 & 89.9 & 95.1 & 91.3 \\
PCA, $k=2048$ & 85.3 & 95.5 & 90.2 & 94.7 & \textbf{84.7} & \textbf{90.6} & \textbf{95.5} & 84.9 & 94.7 & 94.3 & 90.1 & 95.1 & 91.3 \\
PCA, $k=4096$ & 85.6 & 95.5 & 90.2 & 94.7 & 84.4 & 90.1 & \textbf{95.5} & 84.9 & 94.9 & 94.3 & 89.7 & 95.0 & 91.2 \\
Small ensemble & \textbf{86.1} & 95.6 & 90.5 & \textbf{95.4} & 84.5 & 90.2 & \textbf{95.5} & \textbf{85.3} & \textbf{95.2} & \textbf{94.9} & \textbf{90.4} & \textbf{95.3} & \textbf{91.6} \\
\bottomrule
\end{tabular}
\setlength\tabcolsep{6.4 pt}
\begin{tabular}{ l | c c c c c c c | c }
\toprule
\multicolumn{9}{c}{DomainNet (3-shot) No Augmentation} \\
 \midrule
 Method & R $\rightarrow$ C & R $\rightarrow$ P & P $\rightarrow$ C & C $\rightarrow$ S & S $\rightarrow$ P & R $\rightarrow$ S & P $\rightarrow$ R & Mean\\
 \hline
Best ensemble member & 83.0 & 84.0 & 83.0 & 74.5 & 84.2 & 73.9 & 92.1 & 81.7 \\
PCA, $k=1024$ & 83.4 & 84.2 & 83.7 & 75.3 & 84.7 & 74.8 & \textbf{92.6} & 82.7 \\
PCA, $k=2048$ & 83.7 & 84.5 & 84.0 & 75.6 & 84.8 & \textbf{75.1} & 92.5 & 82.9 \\
PCA, $k=4096$ & 83.8 & 84.6 & 84.0 & \textbf{75.7} & 84.9 & \textbf{75.1} & 92.5 & 82.9 \\
PCA, $k=8192$ & 83.9 & 84.6 & 84.0 & 75.6 & 84.7 & 74.8 & 92.5 & 82.9 \\
Small ensemble & \textbf{84.0} & \textbf{84.7} & \textbf{84.2} & 75.2 & \textbf{85.1} & 75.0 & 92.5 & \textbf{83.0} \\
 \bottomrule
\end{tabular}
}
\label{tab:PCA-CORAL} 
\end{table}
\setlength\tabcolsep{6 pt}

\subsection{Hyperparameter Results}
\begin{figure}
\centering
\begin{subfigure}{.49\textwidth}
  \centering
  \includegraphics[width=1.\linewidth]{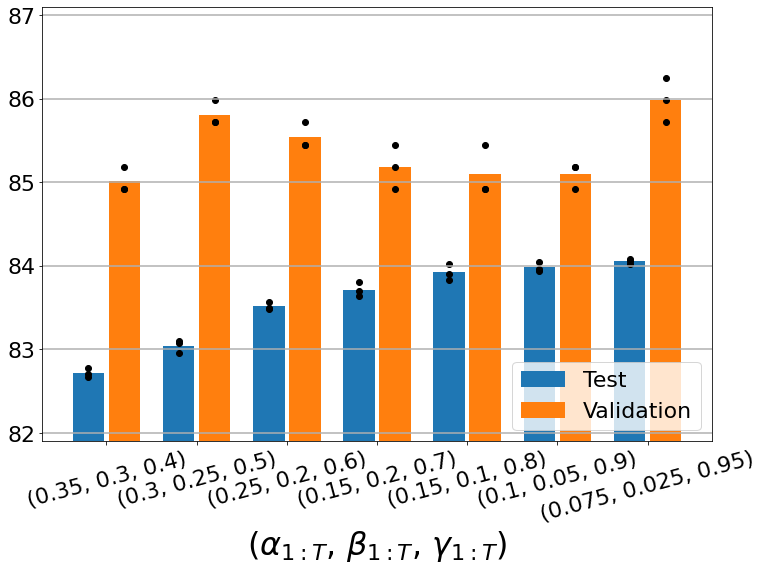}
\end{subfigure}
\begin{subfigure}{.49\textwidth}
  \centering
  \includegraphics[width=1.\linewidth]{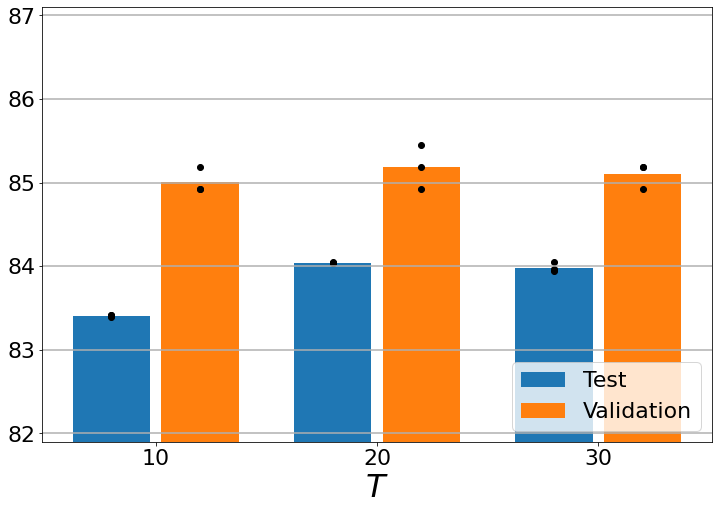}
\end{subfigure}
\begin{subfigure}{.49\textwidth}
  \centering
  \includegraphics[width=1.\linewidth]{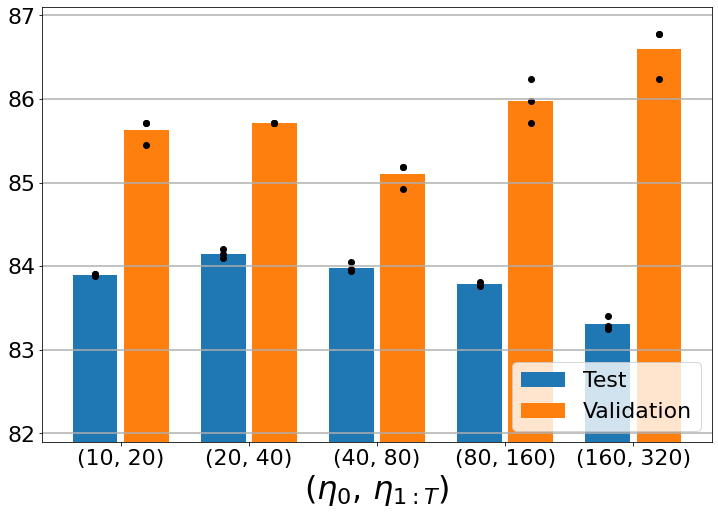}
\end{subfigure}
\begin{subfigure}{.49\textwidth}
  \centering
  \includegraphics[width=1.\linewidth]{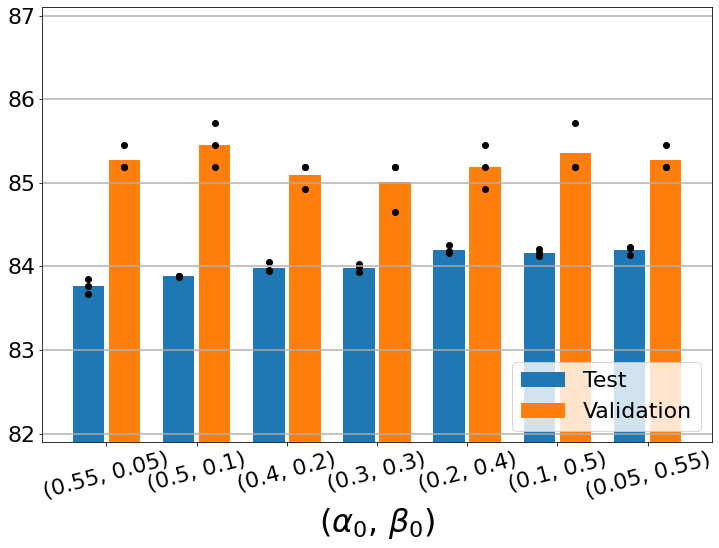}
\end{subfigure}
\begin{subfigure}{.49\textwidth}
  \centering
  \includegraphics[width=1.\linewidth]{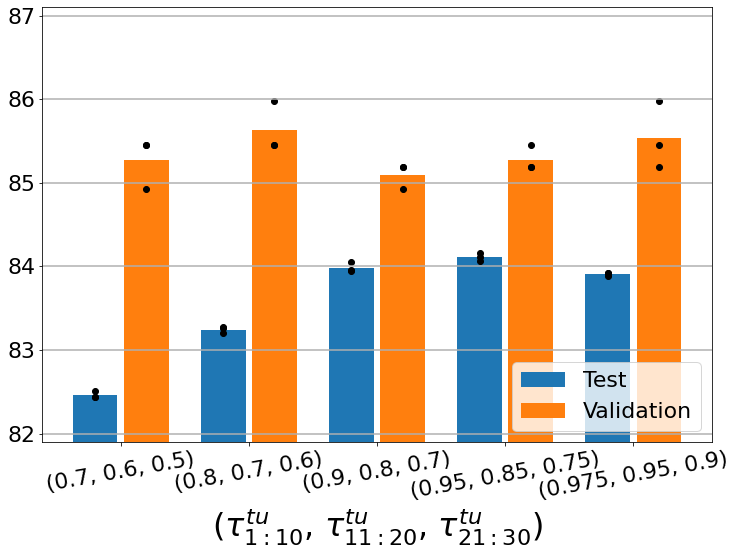}
\end{subfigure}
\begin{subfigure}{.49\textwidth}
  \centering
  \includegraphics[width=1.\linewidth]{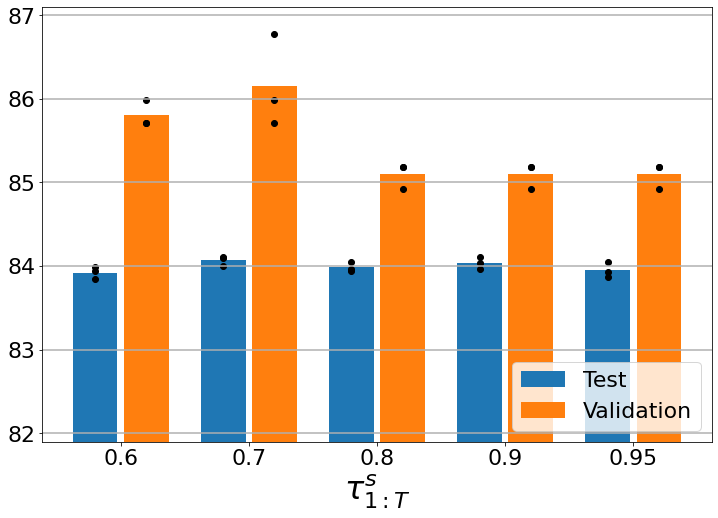}
\end{subfigure}
\caption{DomainNet 3-shot R to C hyperparameter results.}
\label{fig:domainnet-3-shot-hyper}
\end{figure}

\begin{figure}
\centering
\begin{subfigure}{.49\textwidth}
  \centering
  \includegraphics[width=1.\linewidth]{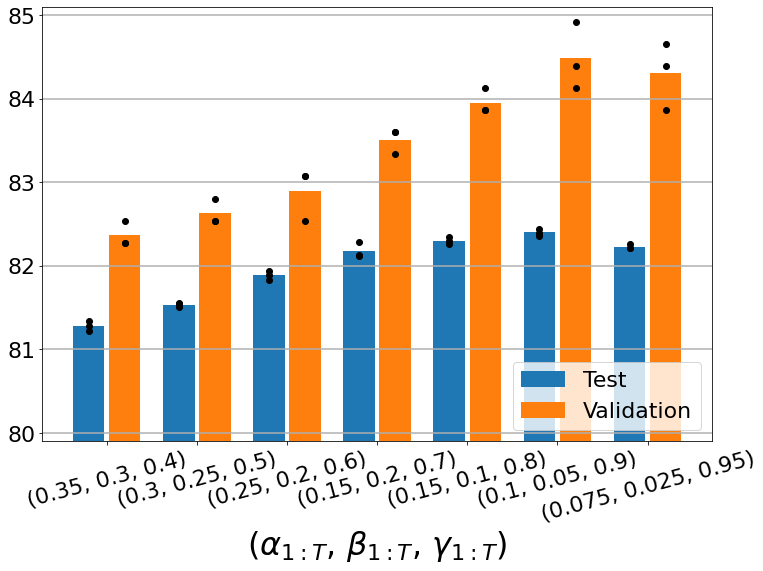}
\end{subfigure}
\begin{subfigure}{.49\textwidth}
  \centering
  \includegraphics[width=1.\linewidth]{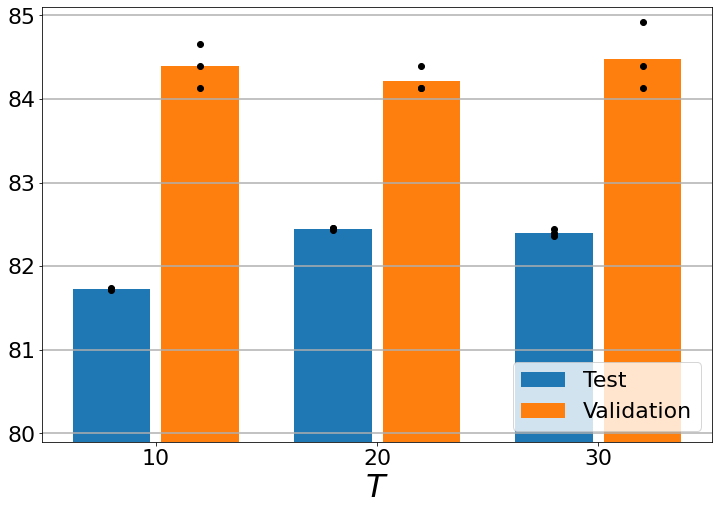}
\end{subfigure}
\begin{subfigure}{.49\textwidth}
  \centering
  \includegraphics[width=1.\linewidth]{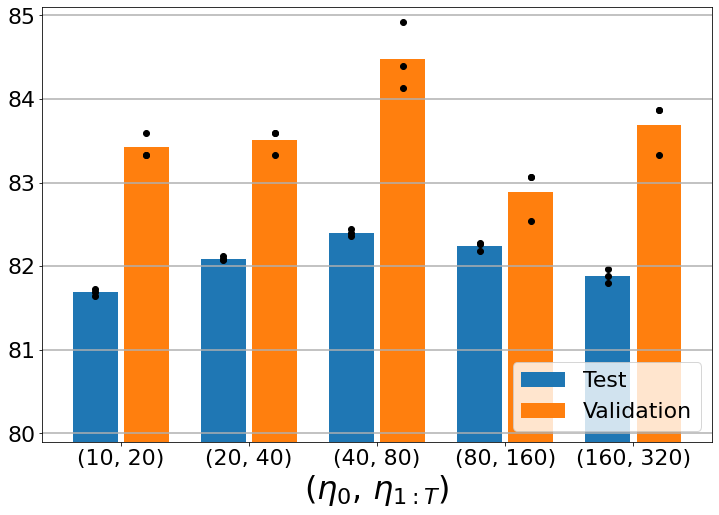}
\end{subfigure}
\begin{subfigure}{.49\textwidth}
  \centering
  \includegraphics[width=1.\linewidth]{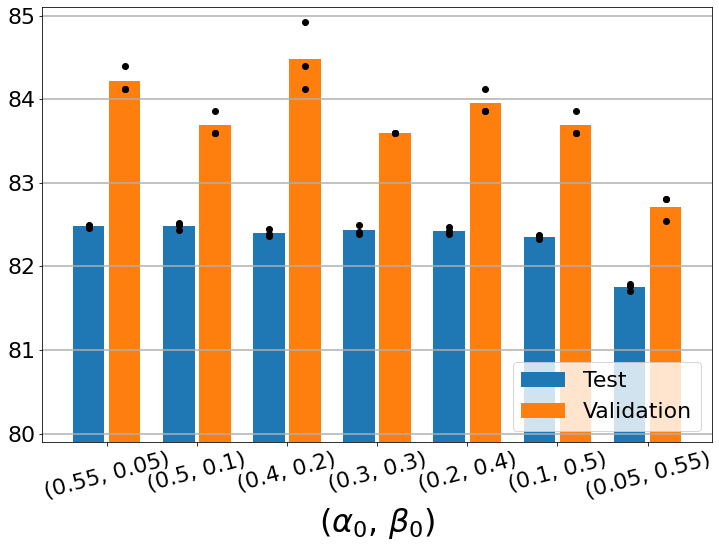}
\end{subfigure}
\begin{subfigure}{.49\textwidth}
  \centering
  \includegraphics[width=1.\linewidth]{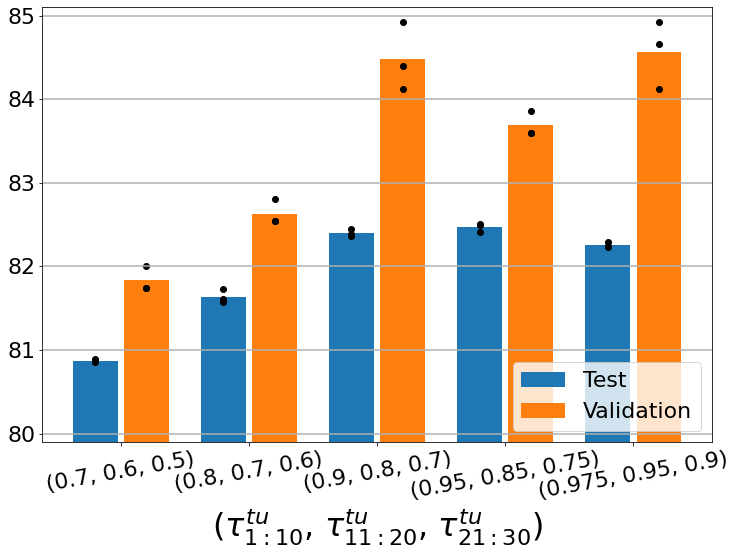}
\end{subfigure}
\begin{subfigure}{.49\textwidth}
  \centering
  \includegraphics[width=1.\linewidth]{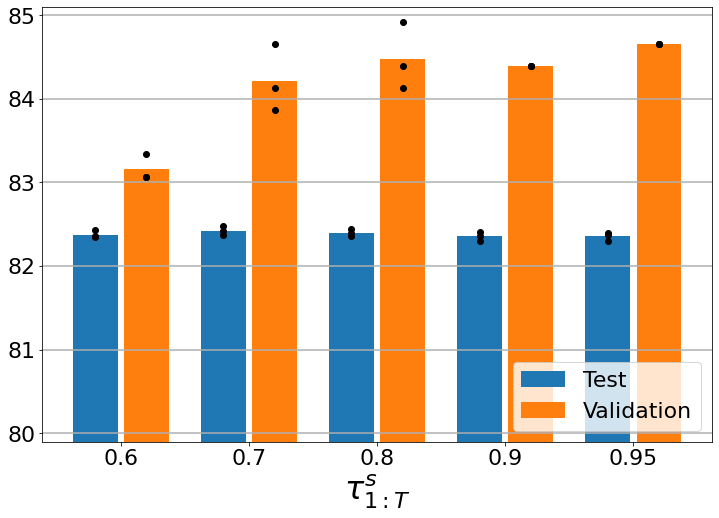}
\end{subfigure}
\caption{DomainNet 1-shot R to C hyperparameter results.}
\label{fig:domainnet-1-shot-hyper}
\end{figure}

\begin{figure}
\centering
\begin{subfigure}{.49\textwidth}
  \centering
  \includegraphics[width=1.\linewidth]{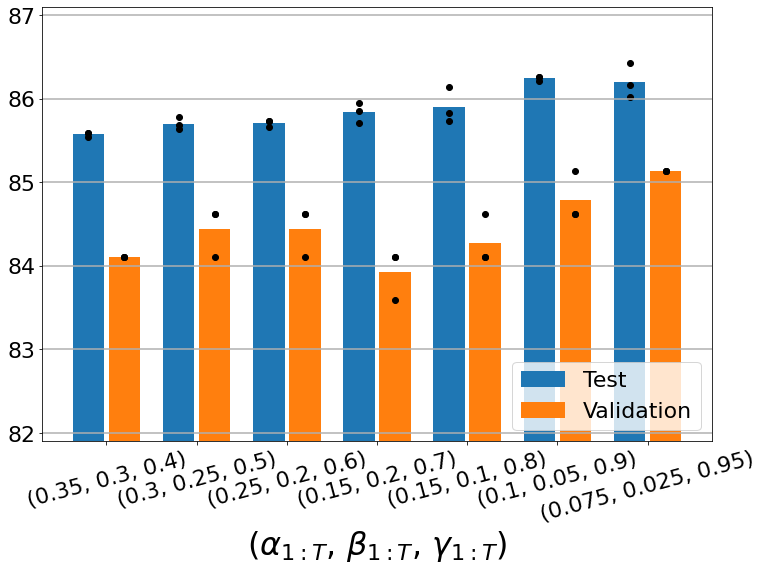}
\end{subfigure}
\begin{subfigure}{.49\textwidth}
  \centering
  \includegraphics[width=1.\linewidth]{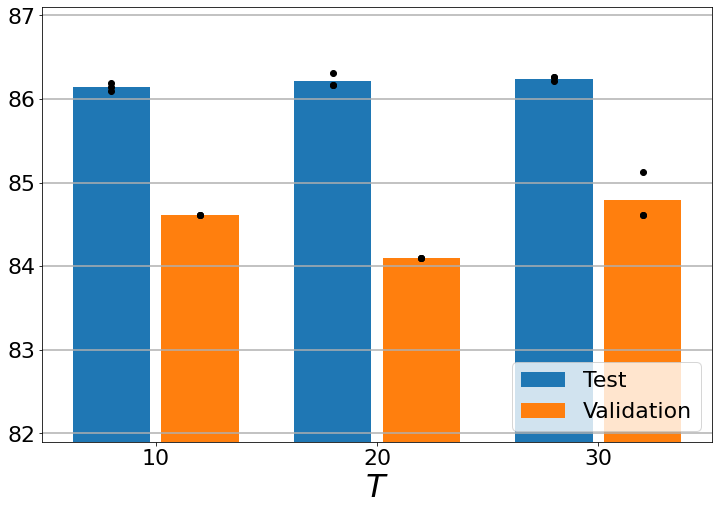}
\end{subfigure}
\begin{subfigure}{.49\textwidth}
  \centering
  \includegraphics[width=1.\linewidth]{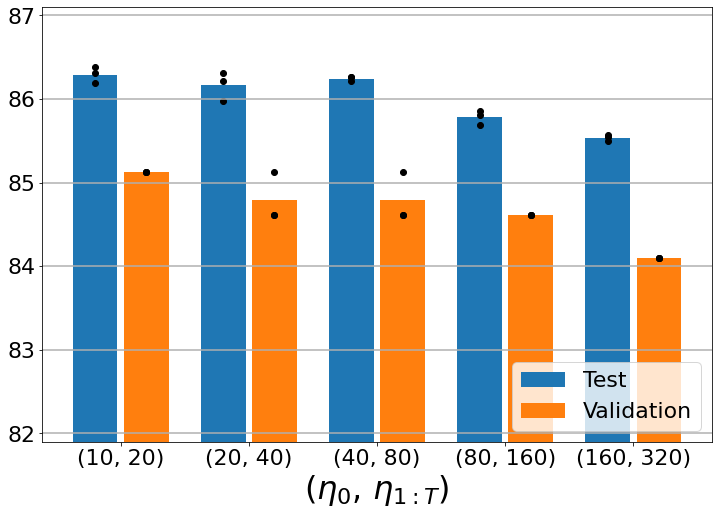}
\end{subfigure}
\begin{subfigure}{.49\textwidth}
  \centering
  \includegraphics[width=1.\linewidth]{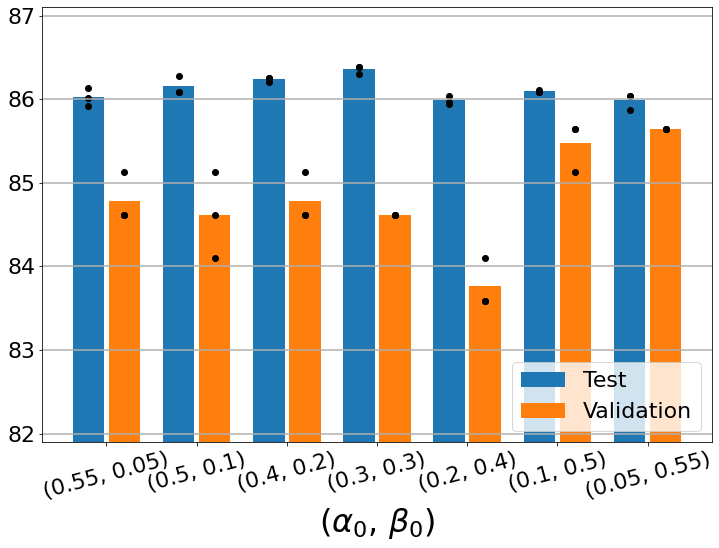}
\end{subfigure}
\begin{subfigure}{.49\textwidth}
  \centering
  \includegraphics[width=1.\linewidth]{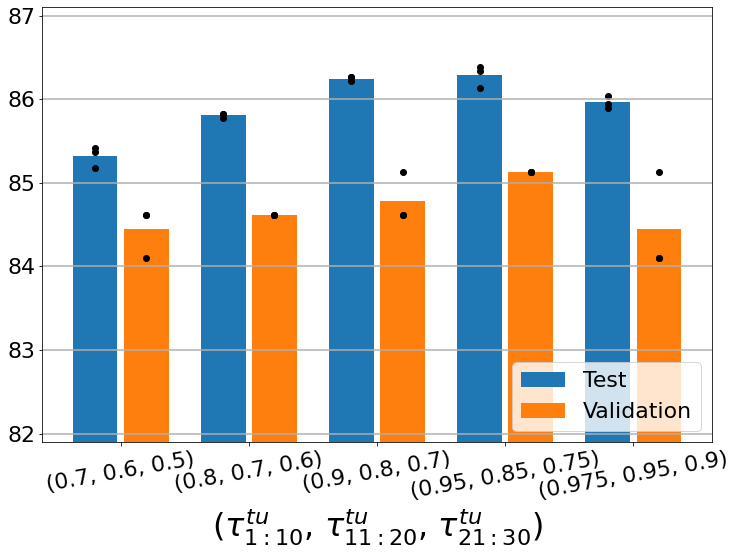}
\end{subfigure}
\begin{subfigure}{.49\textwidth}
  \centering
  \includegraphics[width=1.\linewidth]{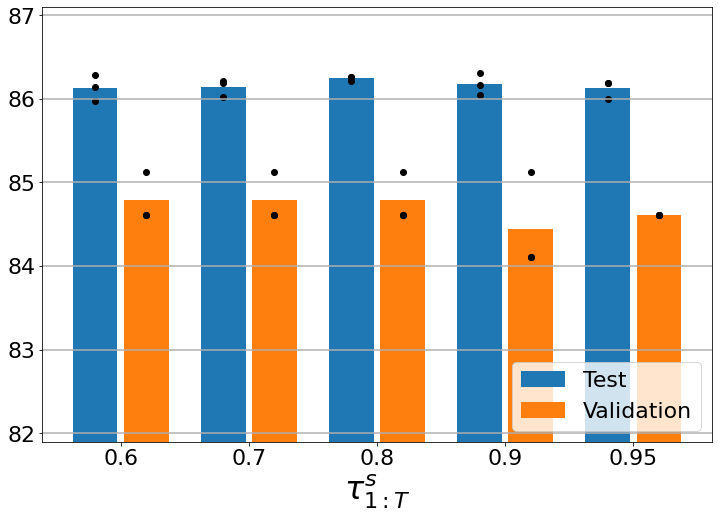}
\end{subfigure}
\caption{Office-Home 3-shot R to C hyperparameter results.}
\label{fig:officehome-3-shot-hyper}
\end{figure}

\begin{figure}
\centering
\begin{subfigure}{.49\textwidth}
  \centering
  \includegraphics[width=1.\linewidth]{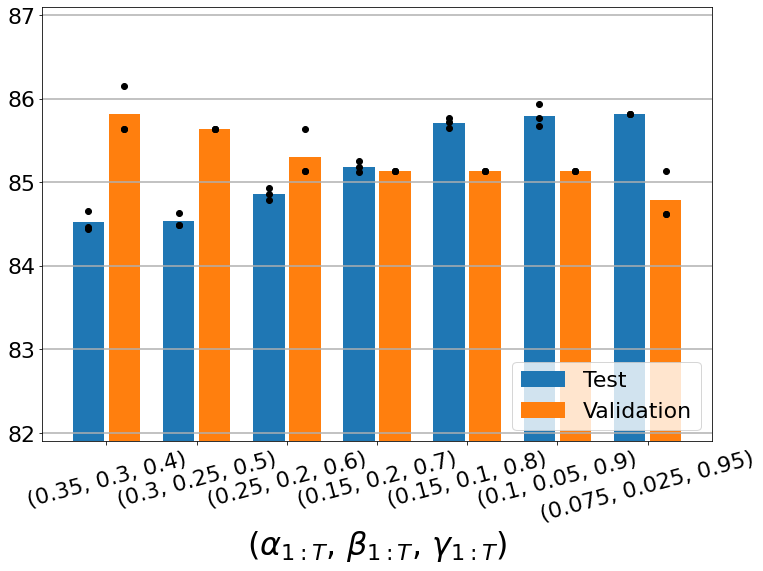}
\end{subfigure}
\begin{subfigure}{.49\textwidth}
  \centering
  \includegraphics[width=1.\linewidth]{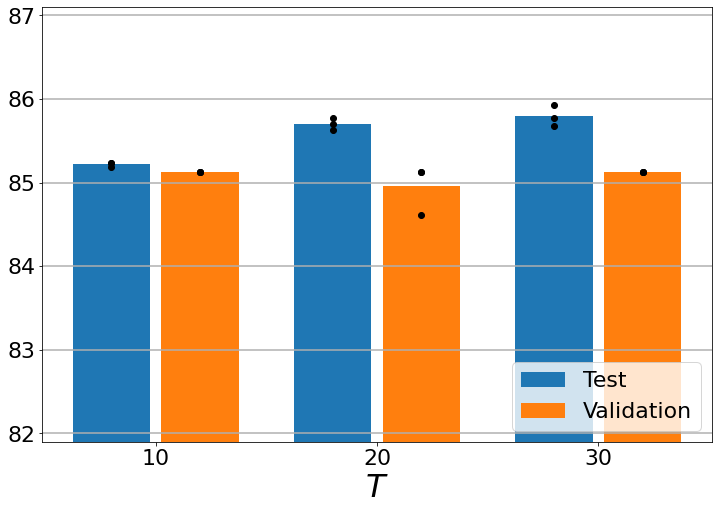}
\end{subfigure}
\begin{subfigure}{.49\textwidth}
  \centering
  \includegraphics[width=1.\linewidth]{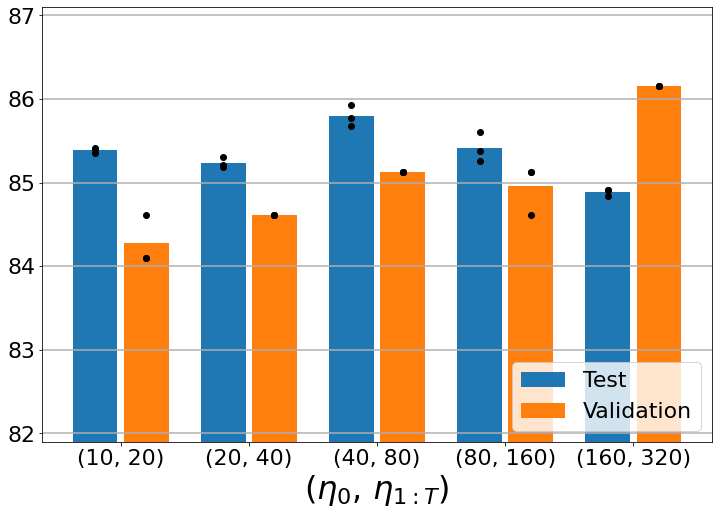}
\end{subfigure}
\begin{subfigure}{.49\textwidth}
  \centering
  \includegraphics[width=1.\linewidth]{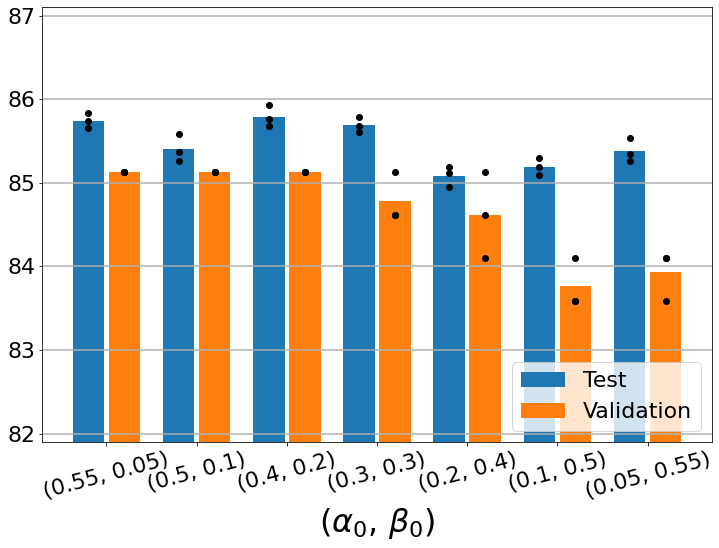}
\end{subfigure}
\begin{subfigure}{.49\textwidth}
  \centering
  \includegraphics[width=1.\linewidth]{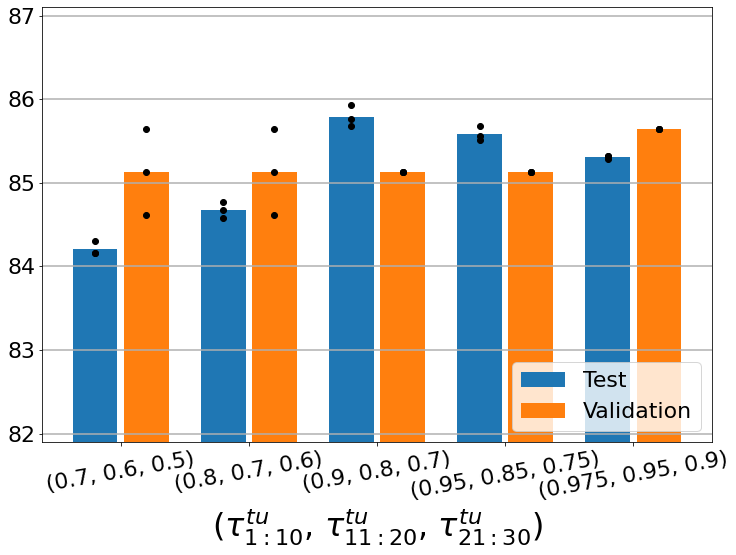}
\end{subfigure}
\begin{subfigure}{.49\textwidth}
  \centering
  \includegraphics[width=1.\linewidth]{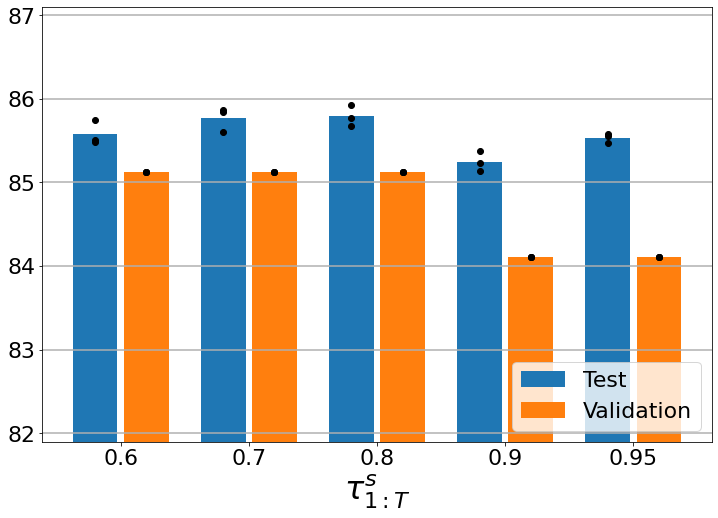}
\end{subfigure}
\caption{Office-Home 1-shot R to C hyperparameter results.}
\label{fig:officehome-1-shot-hyper}
\end{figure}

\begin{figure}
\centering
\begin{subfigure}{.49\textwidth}
  \centering
  \includegraphics[width=1.\linewidth]{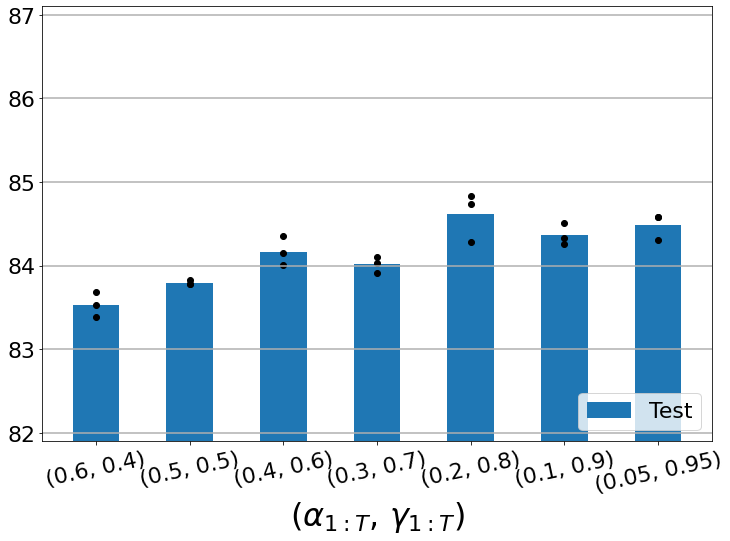}
\end{subfigure}
\begin{subfigure}{.49\textwidth}
  \centering
  \includegraphics[width=1.\linewidth]{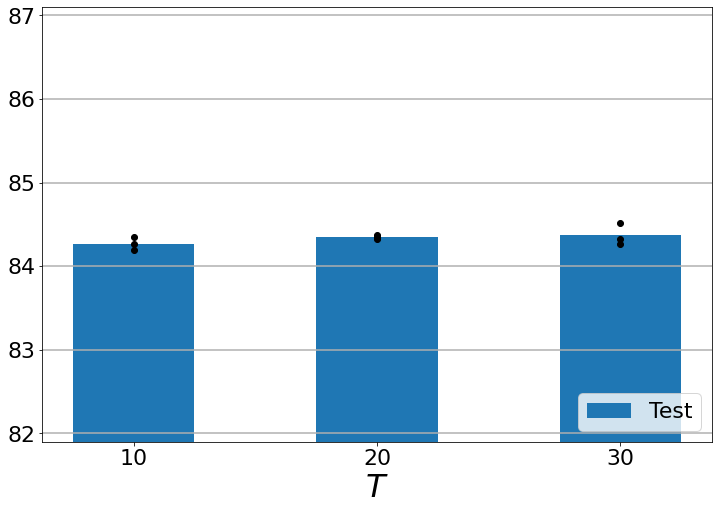}
\end{subfigure}
\begin{subfigure}{.49\textwidth}
  \centering
  \includegraphics[width=1.\linewidth]{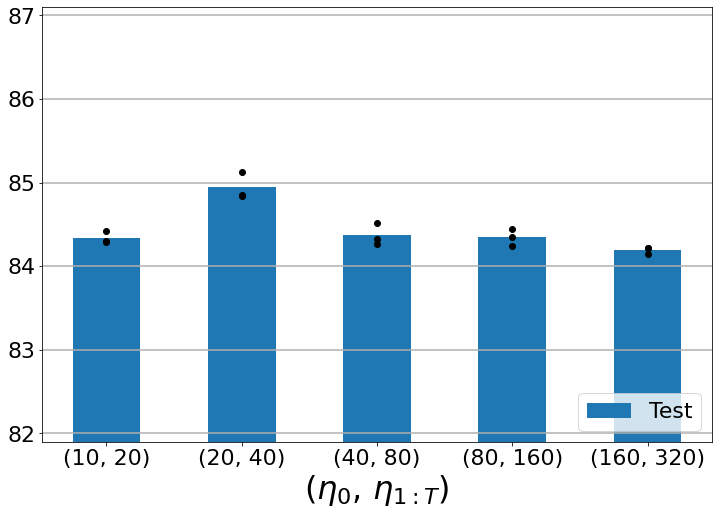}
\end{subfigure}
\begin{subfigure}{.49\textwidth}
  \centering
  \includegraphics[width=1.\linewidth]{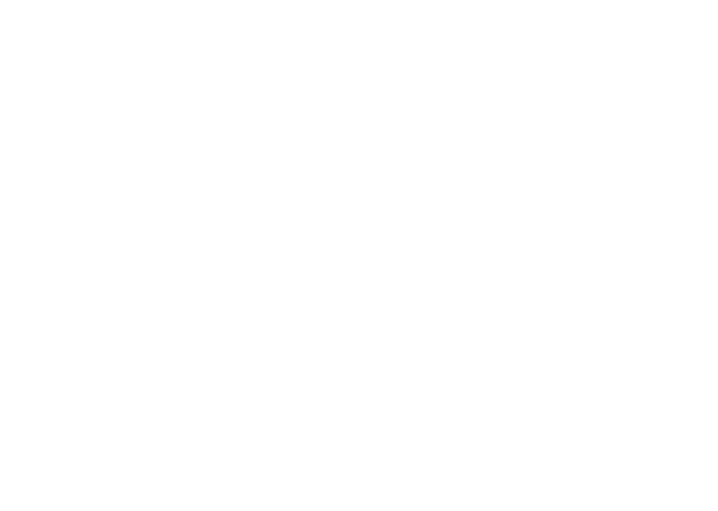}
\end{subfigure}
\begin{subfigure}{.49\textwidth}
  \centering
  \includegraphics[width=1.\linewidth]{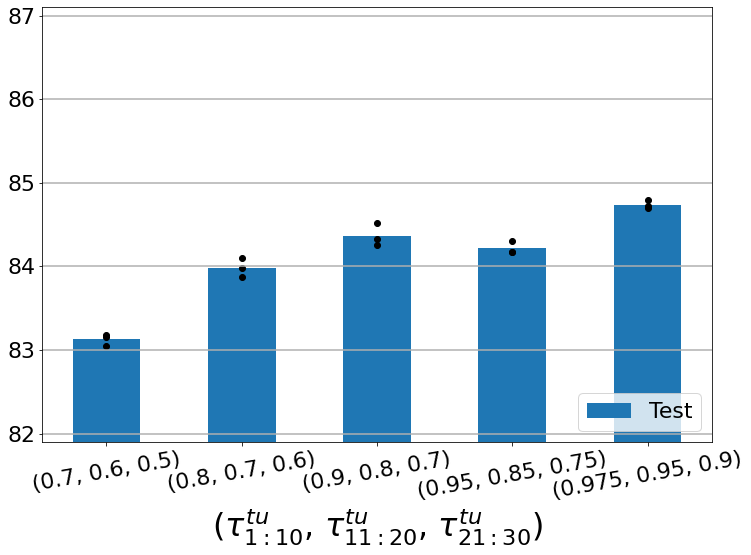}
\end{subfigure}
\begin{subfigure}{.49\textwidth}
  \centering
  \includegraphics[width=1.\linewidth]{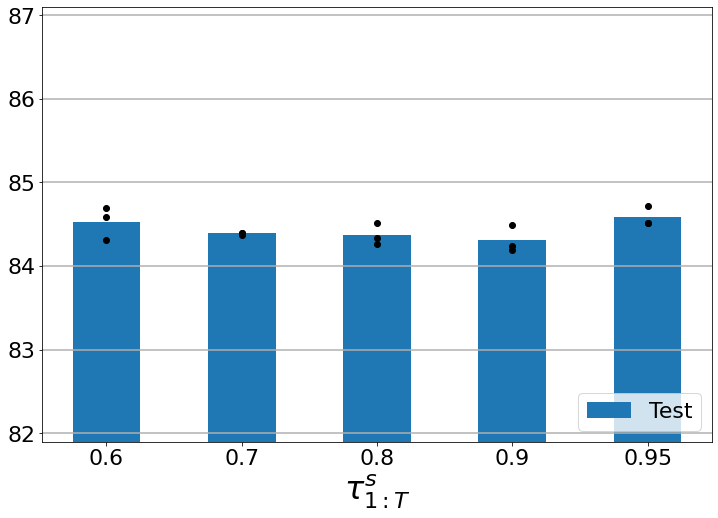}
\end{subfigure}
\caption{Office-Home zero-shot (UDA) R to C hyperparameter results.}
\label{fig:officehome-0-shot-hyper}
\end{figure}

We plot the target accuracy with respect to different hyperparameter settings for 3-shot DomainNet, 1-shot DomainNet, 3-shot Office-Home, 1-shot Office-Home and zero-shot Office-Home in Figures \ref{fig:domainnet-3-shot-hyper}, \ref{fig:domainnet-1-shot-hyper}, \ref{fig:officehome-3-shot-hyper}, \ref{fig:officehome-1-shot-hyper} and \ref{fig:officehome-0-shot-hyper}, respectively. The y-axis is target accuracy for all plots. In the SSDA setting, there is a validation set containing 3 labeled samples from each class; in the UDA setting, there is no validation set. We only show results for the Real to Clipart (R $\rightarrow$ C) task. We run 3 trials for each hyperparameter setting and plot the target accuracy of each individual trial as a black dot. We identify 6 degrees of freedom (5 for UDA); in each figure, we vary the hyperparameter choices along one degree of freedom while holding the remaining hyperparameters constant. The default hyperparameters were stated in Section 4: $\eta_0 = 40$, $\eta_{1:T} = 80$, $T=30$, $\alpha_0 = 0.4$, $\beta_0 = 0.2$, $\alpha_{1:T} = 0.1$, $\beta_{1:T} = 0.05$, $\gamma_{1:T} = 0.9$, $\tau^s_{1:T} = 0.8$, $\tau^{tu}_{1:10} = 0.9$, $\tau^{tu}_{11:20} = 0.8$, $\tau^{tu}_{21:30} = 0.7$.

\paragraph{Observations} During the self-training stage, very little weight should be placed on the source data. Our default choices for $\tau^{tu}_t$ are close to optimal. Our default choices for the learning rates are close to optimal. Running self-training for more than $T=30$ iterations is unlikely to produce better results. Note that the optimal weights between labeled target data and labeled source data ($\alpha_0$ and $\beta_0$) depend on the number of labeled samples in each category (Section 6 in \cite{ben2010theory}). It is not possible to choose a value for $\alpha_0$ and $\beta_0$ that works well for all settings; prior SSDA methods use $\alpha_0 = \beta_0$.

\subsection{Detailed Results on Each Ensemble Member}
We list the unlabeled target accuracies for each ensemble member in Tables \ref{tab:office-home-3-shot-detailed}, \ref{tab:domainnet-3-shot-detailed}, \ref{tab:office-home-1-shot-detailed}, \ref{tab:domainnet-1-shot-detailed} and \ref{tab:office-home-0-shot-detailed}. The seven ensemble members are listed in the same order as Table \ref{tab:ensemble-details}. In the name of the backbone, ``384'' and ``224'' refer to the resolution of the input image; ``1K'' refers to pretraining on ImageNet-22K and fine-tuning on ImageNet-1K; and ``22K'' refers to pretraining on ImageNet-22K with no fine-tuning. In these tables, we report both the ``Majority vote'' and ``Averaged'' predictions. In majority voting, each ensemble member votes for its own prediction, and the label with the most votes wins. Majority voting is usually worse than simply averaging member predictions, because a confident vote is weighted the same as an unconfident vote. The ``Averaged'' row in the ``No Augmentation'' table corresponds to the small ensemble result in the main paper. The ``Averaged'' row in the ``Full Ensemble'' table corresponds to the large ensemble result in the main paper.

\setlength\tabcolsep{1.7 pt}
\begin{table}
\caption{Comparison of different ensembling strategies. }
\centering
{ \small
\begin{tabular}{ l | c c c c c c c c c c c c | c }
\toprule
\multicolumn{14}{c}{Office-Home (3-shot) Small Ensemble No Augmentation} \\
 \midrule
 Method & R - C & R - P & R - A & P - R & P - C & P - A & A - P & A - C & A - R & C - R & C - A & C - P & Mean \\
 \hline
Best ensemble member  & 85.0 & \textbf{95.8} & \textbf{90.6} & 94.9 & 83.3 & 89.7 & \textbf{95.5} & 84.1 & 94.7 & 94.2 & 89.6 & 94.7 & 90.7 \\
Majority vote & 86.2 & 95.6 & 90.3 & \textbf{95.2} & 84.5 & 90.2 & 95.3 & 85.1 & 94.9 & \textbf{94.9} & 90.2 & \textbf{95.3} & 91.5 \\
 {\white j} + Confidence weighting & \textbf{86.3} & 95.6 & 90.5 & \textbf{95.2} & \textbf{84.7} & \textbf{90.5} & 95.4 & 85.2 & 95.0 & 94.8 & 90.3 & 95.2 & \textbf{91.6} \\
 {\white j} + Validation acc weighting & \textbf{86.3} & 95.6 & 90.5 & \textbf{95.2} & 84.6 & \textbf{90.5} & \textbf{95.5} & 85.2 & 95.0 & 94.8 & 90.3 & 95.2 & \textbf{91.6} \\
Averaged & 86.2 & 95.6 & 90.5 & \textbf{95.2} & \textbf{84.7} & \textbf{90.5} & 95.4 & \textbf{85.3} & \textbf{95.1} & 94.8 & \textbf{90.4} & 95.2 & \textbf{91.6} \\
\bottomrule
\end{tabular}
\begin{tabular}{ l | c c c c c c c c c c c c | c }
\toprule
\multicolumn{14}{c}{Office-Home (3-shot) Large Ensemble} \\
 \midrule
 Method & R - C & R - P & R - A & P - R & P - C & P - A & A - P & A - C & A - R & C - R & C - A & C - P & Mean \\
 \hline
Best ensemble member  & 86.1 & \textbf{95.8} & 90.6 & 94.9 & 84.5 & 89.7 & \textbf{95.5} & \textbf{86.2} & 94.7 & 94.2 & 90.1 & 94.7 & 91.4 \\
Majority vote & 87.0 & 95.7 & 90.7 & 95.1 & \textbf{85.0} & \textbf{91.0} & 95.3 & 86.1 & 94.9 & \textbf{95.0} & 91.1 & \textbf{95.3} & 91.8 \\
 {\white j} + Confidence weighting  & \textbf{87.1} & 95.7 & \textbf{90.9} & \textbf{95.3} & 84.9 & \textbf{91.0} & 95.3 & \textbf{86.2} & 94.9 & \textbf{95.0} & 91.0 & 95.2 & \textbf{91.9} \\
 {\white j} + Validation acc weighting& \textbf{87.1} & \textbf{95.8} & 90.8 & \textbf{95.3} & \textbf{85.0} & \textbf{91.0} & 95.3 & \textbf{86.2} & 94.9 & \textbf{95.0} & 91.0 & 95.2 & \textbf{91.9} \\
Averaged & 87.0 & 95.6 & \textbf{90.9} & \textbf{95.3} & 84.9 & 90.9 & 95.4 & 86.1 & \textbf{95.0} & 94.9 & \textbf{91.2} & \textbf{95.3} & \textbf{91.9} \\
\bottomrule
\end{tabular}
\setlength\tabcolsep{5.9 pt}
\begin{tabular}{ l | c c c c c c c | c }
\toprule
\multicolumn{9}{c}{DomainNet (3-shot) Small Ensemble No Augmentation} \\
 \midrule
 Method & R $\rightarrow$ C & R $\rightarrow$ P & P $\rightarrow$ C & C $\rightarrow$ S & S $\rightarrow$ P & R $\rightarrow$ S & P $\rightarrow$ R & Mean\\
 \hline
Best ensemble member  & 83.0 & 84.1 & 82.6 & 74.6 & 84.3 & 73.7 & 92.1 & 81.8 \\
Majority vote & 83.9 & 84.6 & 84.0 & 75.2 & 84.9 & 74.6 & \textbf{92.5} & 82.8 \\
 {\white j} + Confidence weighting  & 83.9 & \textbf{84.8} & 84.1 & 75.2 & 84.9 & 74.9 & \textbf{92.5} & 82.9 \\
 {\white j} + Validation acc weighting& \textbf{84.0} & \textbf{84.8} & 84.1 & \textbf{75.3} & \textbf{85.0} & 74.9 & \textbf{92.5} & 82.9 \\
Averaged & \textbf{84.0} & \textbf{84.8} & \textbf{84.2} & \textbf{75.3} & \textbf{85.0} & \textbf{75.0} & \textbf{92.5} & \textbf{83.0} \\
 \bottomrule
\end{tabular}
\begin{tabular}{ l | c c c c c c c | c }
\toprule
\multicolumn{9}{c}{DomainNet (3-shot) Large Ensemble} \\
 \midrule
 Method & R $\rightarrow$ C & R $\rightarrow$ P & P $\rightarrow$ C & C $\rightarrow$ S & S $\rightarrow$ P & R $\rightarrow$ S & P $\rightarrow$ R & Mean\\
 \hline
Best ensemble member  & 83.0 & 84.2 & 83.1 & 75.2 & 84.3 & 74.7 & 92.1 & 82.4 \\
Majority vote & 84.0 & 84.9 & \textbf{84.5} & 75.8 & 8\textbf{5.2} & 75.2 & 92.4 & \textbf{83.2}\\
 {\white j} + Confidence weighting & 84.0 & \textbf{85.1} & 84.4 & 75.8 & \textbf{85.2} & \textbf{75.4} & \textbf{92.5} & \textbf{83.2}\\
 {\white j} + Validation acc weighting & 84.0 & \textbf{85.1} & \textbf{84.5} & \textbf{75.9} & \textbf{85.2} & \textbf{75.4} & \textbf{92.5} & \textbf{83.2}\\
Averaged & \textbf{84.1} & 85.0 & \textbf{84.5} & \textbf{75.9} & \textbf{85.2} & \textbf{75.4} & \textbf{92.5} & \textbf{83.2} \\
 \bottomrule
\end{tabular}
}
\label{tab:Different-ensembling-strategies} 
\end{table}
\setlength\tabcolsep{6 pt}

\newpage

\newpage

\begin{table}
\caption{Office-Home 3-shot detailed results}
\centering
\setlength\tabcolsep{2.5 pt}
{ \small
\begin{tabular}{ l | c c c c c c c c c c c c | c }
\toprule
\multicolumn{14}{c}{Office-Home 3-shot Grayscale} \\
\midrule
Backbone & R - C & R - P & R - A & P - R & P - C & P - A & A - P & A - C & A - R & C - R & C - A & C - P & Mean\\
\hline
ConvNeXt-384-1K & 81.3 & 95.2 & 89.9 & 93.6 & 80.8 & 88.8 & 94.6 & 81.0 & 93.4 & 93.2 & 88.8 & 93.9 & 89.5 \\
ConvNeXt-224-1K & 82.8 & 93.8 & 88.6 & 93.7 & 80.6 & 86.8 & 92.9 & 81.5 & 93.1 & 93.0 & 88.5 & 93.5 & 89.1 \\
ConvNeXt-224-22K & 81.4 & 94.2 & 88.8 & 94.0 & 80.8 & 87.3 & 93.0 & 81.4 & 93.5 & 92.9 & 87.5 & 93.5 & 89.0 \\
Swin-224-1K & 82.3 & 93.2 & 86.9 & 93.6 & 80.3 & 85.8 & 92.3 & 81.4 & 92.7 & 92.4 & 87.1 & 92.9 & 88.4 \\
Swin-224-22K & 82.2 & 93.5 & 87.4 & 93.7 & 79.1 & 85.3 & 93.0 & 81.4 & 93.2 & 92.6 & 86.1 & 92.0 & 88.3 \\
Swin-384-1K & 82.6 & 93.6 & 87.7 & 93.7 & 81.0 & 86.3 & 93.5 & 82.5 & 93.1 & 93.2 & 87.0 & 93.8 & 89.0 \\
Swin-384-22K & 82.5 & 93.7 & 88.5 & 94.2 & 81.2 & 87.5 & 92.8 & 82.8 & 93.3 & 92.7 & 87.6 & 92.6 & 89.1 \\
\hline
Max over members & 82.8 & 95.2 & 89.9 & 94.2 & 81.2 & 88.8 & 94.6 & 82.8 & 93.5 & 93.2 & 88.8 & 93.9 & 89.5 \\
Majority vote & 83.7 & 94.5 & 89.5 & 94.4 & 82.3 & 88.9 & 93.7 & 83.8 & 94.0 & 93.9 & 89.6 & 94.3 & 90.2 \\
Averaged & 84.0 & 94.5 & 90.0 & 94.5 & 82.4 & 89.0 & 93.8 & 84.0 & 94.1 & 93.9 & 89.6 & 94.2 & 90.3 \\
\bottomrule
\end{tabular}

\begin{tabular}{ l | c c c c c c c c c c c c | c }
\toprule
\multicolumn{14}{c}{Office-Home 3-shot Perspective} \\
\midrule
Backbone & R - C & R - P & R - A & P - R & P - C & P - A & A - P & A - C & A - R & C - R & C - A & C - P & Mean\\
\hline
ConvNeXt-384-1K & 85.2 & 95.7 & 90.0 & 94.4 & 83.5 & 89.7 & 95.1 & 85.8 & 94.3 & 94.1 & 90.4 & 94.5 & 91.0 \\
ConvNeXt-224-1K & 85.2 & 94.8 & 89.5 & 93.4 & 83.5 & 89.0 & 94.1 & 85.7 & 93.6 & 93.9 & 90.1 & 93.6 & 90.5 \\
ConvNeXt-224-22K & 85.4 & 95.4 & 89.1 & 94.3 & 84.2 & 88.5 & 94.5 & 85.1 & 94.1 & 94.0 & 89.1 & 94.4 & 90.7 \\
Swin-224-1K & 84.6 & 94.6 & 88.4 & 93.8 & 83.9 & 87.9 & 93.4 & 84.7 & 93.9 & 93.8 & 88.0 & 92.8 & 90.0 \\
Swin-224-22K & 84.8 & 95.2 & 88.6 & 94.1 & 82.9 & 88.0 & 93.9 & 83.8 & 94.0 & 93.5 & 88.6 & 93.9 & 90.1 \\
Swin-384-1K & 86.1 & 95.1 & 89.3 & 94.4 & 84.0 & 88.8 & 94.8 & 85.7 & 94.2 & 94.0 & 89.2 & 94.4 & 90.8 \\
Swin-384-22K & 85.4 & 95.3 & 89.7 & 94.7 & 83.1 & 89.1 & 94.9 & 85.3 & 94.3 & 94.3 & 89.6 & 94.3 & 90.8 \\
\hline
Max over members & 86.1 & 95.7 & 90.0 & 94.7 & 84.2 & 89.7 & 95.1 & 85.8 & 94.3 & 94.3 & 90.4 & 94.5 & 91.0 \\
Majority vote & 87.4 & 95.6 & 90.5 & 94.9 & 85.0 & 90.0 & 95.2 & 86.5 & 94.7 & 94.8 & 90.6 & 95.1 & 91.7 \\
Averaged & 87.5 & 95.6 & 90.7 & 95.0 & 85.1 & 90.3 & 95.1 & 86.8 & 94.7 & 94.7 & 91.0 & 95.1 & 91.8 \\
\bottomrule
\end{tabular}

\begin{tabular}{ l | c c c c c c c c c c c c | c }
\toprule
\multicolumn{14}{c}{Office-Home 3-shot RandAugment} \\
\midrule
Backbone & R - C & R - P & R - A & P - R & P - C & P - A & A - P & A - C & A - R & C - R & C - A & C - P & Mean\\
\hline
ConvNeXt-384-1K & 83.0 & 94.9 & 90.3 & 94.4 & 81.5 & 89.2 & 94.9 & 82.4 & 93.8 & 93.2 & 89.5 & 94.2 & 90.1 \\
ConvNeXt-224-1K & 83.8 & 94.6 & 90.1 & 93.7 & 82.6 & 88.5 & 93.6 & 83.3 & 93.3 & 93.4 & 88.2 & 94.0 & 89.9 \\
ConvNeXt-224-22K & 83.7 & 94.7 & 89.0 & 94.3 & 81.9 & 87.9 & 94.2 & 82.7 & 94.1 & 93.3 & 87.6 & 94.3 & 89.8 \\
Swin-224-1K & 82.4 & 93.9 & 88.1 & 94.2 & 81.4 & 87.6 & 94.1 & 82.0 & 93.6 & 93.3 & 88.2 & 92.8 & 89.3 \\
Swin-224-22K & 82.6 & 93.9 & 87.9 & 93.6 & 81.3 & 87.3 & 94.1 & 82.1 & 93.6 & 93.1 & 87.5 & 92.8 & 89.2 \\
Swin-384-1K & 83.8 & 94.8 & 89.0 & 94.4 & 82.5 & 88.8 & 94.1 & 83.4 & 93.9 & 93.5 & 89.3 & 94.5 & 90.2 \\
Swin-384-22K & 84.1 & 95.0 & 89.4 & 94.7 & 82.5 & 88.7 & 94.5 & 83.3 & 94.1 & 93.9 & 88.2 & 94.4 & 90.2 \\
\hline
Max over members & 84.1 & 95.0 & 90.3 & 94.7 & 82.6 & 89.2 & 94.9 & 83.4 & 94.1 & 93.9 & 89.5 & 94.5 & 90.2 \\
Majority vote & 85.9 & 95.4 & 90.8 & 95.1 & 84.8 & 90.3 & 95.1 & 84.8 & 94.9 & 94.9 & 90.0 & 95.0 & 91.4 \\
Averaged & 86.0 & 95.5 & 90.9 & 95.1 & 84.5 & 90.8 & 95.1 & 85.3 & 95.0 & 95.0 & 90.1 & 95.1 & 91.5 \\
\bottomrule
\end{tabular}

\begin{tabular}{ l | c c c c c c c c c c c c | c }
\toprule
\multicolumn{14}{c}{Office-Home 3-shot No Augmentation} \\
\midrule
Backbone & R - C & R - P & R - A & P - R & P - C & P - A & A - P & A - C & A - R & C - R & C - A & C - P & Mean\\
\hline
ConvNeXt-384-1K & 83.9 & 95.8 & 90.6 & 94.3 & 82.0 & 89.3 & 95.4 & 83.3 & 94.3 & 94.0 & 89.6 & 94.7 & 90.6 \\
ConvNeXt-224-1K & 84.1 & 95.0 & 89.8 & 94.2 & 83.0 & 88.9 & 94.6 & 84.1 & 94.0 & 93.7 & 88.8 & 94.6 & 90.4 \\
ConvNeXt-224-22K & 84.8 & 95.5 & 89.4 & 94.6 & 82.4 & 88.8 & 95.2 & 83.3 & 94.2 & 93.8 & 88.3 & 94.6 & 90.4 \\
Swin-224-1K & 83.2 & 94.6 & 88.0 & 94.5 & 83.2 & 88.4 & 94.2 & 83.7 & 94.0 & 94.1 & 87.5 & 93.8 & 89.9 \\
Swin-224-22K & 84.3 & 95.1 & 88.0 & 94.5 & 82.6 & 87.8 & 94.7 & 82.6 & 94.6 & 93.8 & 88.2 & 93.8 & 90.0 \\
Swin-384-1K & 84.5 & 95.0 & 89.0 & 94.5 & 83.0 & 89.4 & 94.3 & 83.7 & 94.1 & 94.0 & 88.4 & 94.6 & 90.4 \\
Swin-384-22K & 85.0 & 95.0 & 89.6 & 94.9 & 82.5 & 89.2 & 94.6 & 84.0 & 94.6 & 94.1 & 88.8 & 94.1 & 90.5 \\
\hline
Max over members & 85.0 & 95.8 & 90.6 & 94.9 & 83.2 & 89.4 & 95.4 & 84.1 & 94.6 & 94.1 & 89.6 & 94.7 & 90.6 \\
Majority vote & 85.9 & 95.5 & 90.3 & 95.3 & 84.5 & 90.3 & 95.4 & 85.3 & 95.1 & 94.8 & 90.1 & 95.3 & 91.5 \\
Averaged & 86.1 & 95.6 & 90.5 & 95.4 & 84.5 & 90.2 & 95.5 & 85.3 & 95.2 & 94.9 & 90.4 & 95.3 & 91.6 \\
\bottomrule
\end{tabular}

\begin{tabular}{ l | c c c c c c c c c c c c | c }
\toprule
\multicolumn{14}{c}{Office-Home 3-shot Full Ensemble} \\
\midrule
Backbone & R - C & R - P & R - A & P - R & P - C & P - A & A - P & A - C & A - R & C - R & C - A & C - P & Mean\\
\hline
Majority vote  {\white jjjjjj } & 86.9 & 95.7 & 90.8 & 95.2 & 85.0 & 90.7 & 95.4 & 86.1 & 94.9 & 94.9 & 91.3 & 95.4 & 91.9 \\
Averaged & 87.0 & 95.7 & 90.8 & 95.1 & 85.0 & 90.7 & 95.3 & 86.3 & 94.9 & 94.9 & 91.2 & 95.3 & 91.9 \\
\bottomrule
\end{tabular}
}
\label{tab:office-home-3-shot-detailed}
\end{table}
\setlength\tabcolsep{6 pt}

\begin{table}
\caption{DomainNet 3-shot detailed results}
\centering
\setlength\tabcolsep{4.5 pt}
{ \small
\begin{tabular}{ l | c c c c c c c | c }
\toprule

\multicolumn{9}{c}{DomainNet 3-shot Grayscale} \\
\midrule
Backbone & R $\rightarrow$ C & R $\rightarrow$ P & P $\rightarrow$ C & C $\rightarrow$ S & S $\rightarrow$ P & R $\rightarrow$ S & P $\rightarrow$ R & Mean\\
\hline
ConvNeXt-384-1K & 79.6 & 82.2 & 79.3 & 75.8 & 83.0 & 74.4 & 90.3 & 80.7 \\
ConvNeXt-224-1K & 78.0 & 80.9 & 78.5 & 74.7 & 80.9 & 74.2 & 89.4 & 79.5 \\
ConvNeXt-224-22K & 79.2 & 80.7 & 78.3 & 73.4 & 80.2 & 71.9 & 89.9 & 79.1 \\
Swin-224-1K & 77.8 & 79.1 & 77.2 & 72.5 & 79.4 & 71.1 & 88.8 & 78.0 \\
Swin-224-22K & 77.8 & 79.7 & 77.5 & 71.7 & 79.7 & 70.3 & 89.3 & 78.0 \\
Swin-384-1K & 78.2 & 80.8 & 79.3 & 73.5 & 81.9 & 72.8 & 89.6 & 79.5 \\
Swin-384-22K & 79.5 & 81.5 & 79.5 & 72.8 & 82.1 & 72.2 & 90.2 & 79.7 \\
\hline
Max over members & 79.6 & 82.2 & 79.5 & 75.8 & 83.0 & 74.4 & 90.3 & 80.7 \\
Majority vote & 81.1 & 82.4 & 80.9 & 76.0 & 83.1 & 74.7 & 90.6 & 81.2 \\
Averaged & 81.3 & 82.6 & 81.2 & 76.0 & 83.2 & 75.0 & 90.6 & 81.4 \\
\bottomrule
\end{tabular}

\begin{tabular}{ l | c c c c c c c | c }
\toprule

\multicolumn{9}{c}{DomainNet 3-shot Perspective} \\
\midrule
Backbone & R $\rightarrow$ C & R $\rightarrow$ P & P $\rightarrow$ C & C $\rightarrow$ S & S $\rightarrow$ P & R $\rightarrow$ S & P $\rightarrow$ R & Mean\\
\hline
ConvNeXt-384-1K & 81.6 & 84.0 & 81.7 & 74.2 & 84.4 & 74.1 & 92.1 & 81.7 \\
ConvNeXt-224-1K & 80.8 & 82.2 & 81.2 & 73.4 & 82.8 & 72.8 & 91.5 & 80.7 \\
ConvNeXt-224-22K & 81.4 & 82.0 & 81.9 & 72.2 & 81.9 & 70.8 & 91.7 & 80.3 \\
Swin-224-1K & 81.0 & 81.4 & 81.8 & 71.0 & 81.8 & 69.7 & 91.0 & 79.7 \\
Swin-224-22K & 81.7 & 81.9 & 81.7 & 70.3 & 81.4 & 69.4 & 91.1 & 79.7 \\
Swin-384-1K & 82.0 & 82.5 & 82.4 & 72.1 & 83.5 & 71.9 & 91.7 & 80.9 \\
Swin-384-22K & 82.2 & 83.1 & 83.1 & 71.9 & 83.3 & 71.5 & 92.0 & 81.0 \\
\hline
Max over members & 82.2 & 84.0 & 83.1 & 74.2 & 84.4 & 74.1 & 92.1 & 81.7 \\
Majority vote & 83.4 & 84.0 & 84.4 & 74.6 & 84.8 & 73.8 & 92.3 & 82.5 \\
Averaged & 83.7 & 84.3 & 84.5 & 74.7 & 84.8 & 74.1 & 92.3 & 82.6 \\
\bottomrule
\end{tabular}

\begin{tabular}{ l | c c c c c c c | c }
\toprule

\multicolumn{9}{c}{DomainNet 3-shot RandAugment} \\
\midrule
Backbone & R $\rightarrow$ C & R $\rightarrow$ P & P $\rightarrow$ C & C $\rightarrow$ S & S $\rightarrow$ P & R $\rightarrow$ S & P $\rightarrow$ R & Mean\\
\hline
ConvNeXt-384-1K & 80.8 & 83.8 & 81.3 & 74.1 & 83.9 & 73.3 & 92.0 & 81.3 \\
ConvNeXt-224-1K & 80.8 & 82.1 & 80.8 & 73.0 & 82.4 & 72.4 & 91.5 & 80.4 \\
ConvNeXt-224-22K & 81.3 & 82.1 & 81.2 & 71.4 & 82.0 & 70.3 & 91.7 & 80.0 \\
Swin-224-1K & 80.3 & 81.2 & 80.4 & 70.7 & 81.2 & 69.9 & 91.3 & 79.3 \\
Swin-224-22K & 80.9 & 81.8 & 80.6 & 69.6 & 81.4 & 69.2 & 91.4 & 79.3 \\
Swin-384-1K & 82.2 & 82.9 & 81.7 & 72.2 & 83.3 & 71.7 & 91.7 & 80.8 \\
Swin-384-22K & 82.5 & 83.3 & 82.8 & 71.1 & 83.4 & 71.4 & 91.8 & 80.9 \\
\hline
Max over members & 82.5 & 83.8 & 82.8 & 74.1 & 83.9 & 73.3 & 92.0 & 81.3 \\
Majority vote & 83.9 & 84.4 & 84.2 & 75.1 & 84.9 & 74.5 & 92.4 & 82.8 \\
Averaged & 84.1 & 84.6 & 84.4 & 75.2 & 85.0 & 74.9 & 92.5 & 82.9 \\
\bottomrule
\end{tabular}

\begin{tabular}{ l | c c c c c c c | c }
\toprule

\multicolumn{9}{c}{DomainNet 3-shot No Augmentation} \\
\midrule
Backbone & R $\rightarrow$ C & R $\rightarrow$ P & P $\rightarrow$ C & C $\rightarrow$ S & S $\rightarrow$ P & R $\rightarrow$ S & P $\rightarrow$ R & Mean\\
\hline
ConvNeXt-384-1K & 81.8 & 84.0 & 81.6 & 74.5 & 84.2 & 73.9 & 92.1 & 81.7 \\
ConvNeXt-224-1K & 81.4 & 82.8 & 81.4 & 73.9 & 83.2 & 73.0 & 91.8 & 81.1 \\
ConvNeXt-224-22K & 82.4 & 82.7 & 82.1 & 73.0 & 82.7 & 71.2 & 91.8 & 80.9 \\
Swin-224-1K & 81.5 & 81.8 & 81.3 & 71.9 & 82.2 & 71.0 & 91.5 & 80.2 \\
Swin-224-22K & 81.9 & 82.5 & 82.0 & 71.0 & 81.9 & 70.3 & 91.5 & 80.2 \\
Swin-384-1K & 82.4 & 83.3 & 82.6 & 72.0 & 84.0 & 72.6 & 91.9 & 81.3 \\
Swin-384-22K & 83.0 & 83.7 & 83.0 & 72.4 & 83.7 & 72.1 & 92.0 & 81.4 \\
\hline
Max over members & 83.0 & 84.0 & 83.0 & 74.5 & 84.2 & 73.9 & 92.1 & 81.7 \\
Majority vote & 83.9 & 84.5 & 84.1 & 75.1 & 85.0 & 74.7 & 92.5 & 82.8 \\
Averaged & 84.0 & 84.7 & 84.2 & 75.2 & 85.1 & 75.0 & 92.5 & 83.0 \\
\bottomrule
\end{tabular}

\begin{tabular}{ l | c c c c c c c | c }
\toprule

\multicolumn{9}{c}{DomainNet 3-shot Full Ensemble} \\
\midrule
Backbone & R $\rightarrow$ C & R $\rightarrow$ P & P $\rightarrow$ C & C $\rightarrow$ S & S $\rightarrow$ P & R $\rightarrow$ S & P $\rightarrow$ R & Mean\\
\hline
Majority vote {\white jjjjjj } & 84.1 & 84.8 & 84.5 & 76.0 & 85.3 & 75.3 & 92.5 & 83.2 \\
Averaged & 84.2 & 84.9 & 84.5 & 76.0 & 85.3 & 75.4 & 92.5 & 83.3 \\
\bottomrule
\end{tabular}
}
\label{tab:domainnet-3-shot-detailed}
\end{table}
\setlength\tabcolsep{6 pt}

\begin{table}
\caption{Office-Home 1-shot detailed results}
\centering
\setlength\tabcolsep{2.5 pt}
{ \small
\begin{tabular}{ l | c c c c c c c c c c c c | c }
\toprule
\multicolumn{14}{c}{Office-Home 1-shot Grayscale} \\
\midrule
Backbone & R - C & R - P & R - A & P - R & P - C & P - A & A - P & A - C & A - R & C - R & C - A & C - P & Mean\\
\hline
ConvNeXt-384-1K & 80.3 & 94.7 & 89.2 & 93.6 & 78.0 & 88.4 & 94.5 & 79.6 & 92.8 & 91.8 & 89.5 & 93.9 & 88.9 \\
ConvNeXt-224-1K & 80.2 & 93.3 & 88.3 & 93.9 & 77.7 & 87.7 & 92.4 & 79.7 & 92.3 & 92.5 & 88.1 & 92.5 & 88.2 \\
ConvNeXt-224-22K & 79.8 & 93.6 & 88.4 & 93.8 & 77.7 & 86.5 & 92.2 & 80.5 & 92.8 & 92.9 & 87.0 & 92.1 & 88.1 \\
Swin-224-1K & 80.0 & 92.8 & 86.7 & 92.7 & 77.2 & 86.2 & 92.2 & 79.5 & 92.1 & 92.7 & 86.0 & 91.3 & 87.4 \\
Swin-224-22K & 80.3 & 93.1 & 87.0 & 92.7 & 75.7 & 85.7 & 93.2 & 79.7 & 92.1 & 92.0 & 85.6 & 91.5 & 87.4 \\
Swin-384-1K & 81.2 & 93.5 & 88.2 & 93.3 & 78.8 & 86.7 & 92.2 & 80.1 & 92.2 & 93.0 & 87.5 & 92.0 & 88.2 \\
Swin-384-22K & 80.9 & 93.5 & 88.7 & 93.0 & 77.0 & 86.5 & 93.0 & 81.7 & 92.1 & 92.5 & 86.8 & 91.7 & 88.1 \\
\hline
Max over members & 81.2 & 94.7 & 89.2 & 93.9 & 78.8 & 88.4 & 94.5 & 81.7 & 92.8 & 93.0 & 89.5 & 93.9 & 88.9 \\
Majority vote & 82.2 & 94.0 & 89.1 & 93.8 & 79.2 & 88.8 & 94.0 & 82.3 & 93.2 & 93.5 & 89.1 & 93.0 & 89.3 \\
Averaged & 82.6 & 94.1 & 89.2 & 93.8 & 79.4 & 88.8 & 94.1 & 82.3 & 93.3 & 93.5 & 89.5 & 93.3 & 89.5 \\
\bottomrule
\end{tabular}

\begin{tabular}{ l | c c c c c c c c c c c c | c }
\toprule
\multicolumn{14}{c}{Office-Home 1-shot Perspective} \\
\midrule
Backbone & R - C & R - P & R - A & P - R & P - C & P - A & A - P & A - C & A - R & C - R & C - A & C - P & Mean\\
\hline
ConvNeXt-384-1K & 84.3 & 95.3 & 90.2 & 94.1 & 81.7 & 89.7 & 94.8 & 84.2 & 93.8 & 93.7 & 90.6 & 94.4 & 90.6 \\
ConvNeXt-224-1K & 84.9 & 94.4 & 89.5 & 93.5 & 82.0 & 88.4 & 93.6 & 85.0 & 93.2 & 93.5 & 89.6 & 93.3 & 90.1 \\
ConvNeXt-224-22K & 84.2 & 95.1 & 89.0 & 94.2 & 81.8 & 88.2 & 94.5 & 84.2 & 93.6 & 93.6 & 89.1 & 93.1 & 90.0 \\
Swin-224-1K & 85.0 & 94.8 & 87.7 & 93.8 & 82.2 & 87.5 & 93.2 & 84.5 & 93.4 & 92.8 & 88.2 & 92.3 & 89.6 \\
Swin-224-22K & 83.7 & 94.5 & 88.4 & 93.6 & 81.7 & 87.6 & 93.6 & 84.0 & 93.2 & 92.8 & 87.9 & 92.9 & 89.5 \\
Swin-384-1K & 85.0 & 94.7 & 89.1 & 93.8 & 83.2 & 88.4 & 93.8 & 86.4 & 93.4 & 94.0 & 89.0 & 93.0 & 90.3 \\
Swin-384-22K & 85.3 & 94.0 & 89.5 & 94.1 & 82.2 & 88.8 & 93.9 & 84.7 & 93.4 & 93.6 & 89.5 & 93.3 & 90.2 \\
\hline
Max over members & 85.3 & 95.3 & 90.2 & 94.2 & 83.2 & 89.7 & 94.8 & 86.4 & 93.8 & 94.0 & 90.6 & 94.4 & 90.6 \\
Majority vote & 86.4 & 95.3 & 90.5 & 94.4 & 84.0 & 90.2 & 94.7 & 86.6 & 94.1 & 94.4 & 90.8 & 94.6 & 91.3 \\
Averaged & 86.8 & 95.3 & 90.3 & 94.5 & 84.1 & 90.4 & 94.8 & 86.6 & 94.2 & 94.4 & 90.9 & 94.6 & 91.4 \\
\bottomrule
\end{tabular}

\begin{tabular}{ l | c c c c c c c c c c c c | c }
\toprule
\multicolumn{14}{c}{Office-Home 1-shot RandAugment} \\
\midrule
Backbone & R - C & R - P & R - A & P - R & P - C & P - A & A - P & A - C & A - R & C - R & C - A & C - P & Mean\\
\hline
ConvNeXt-384-1K & 81.1 & 94.5 & 89.6 & 93.9 & 78.8 & 89.0 & 94.0 & 80.6 & 93.1 & 92.7 & 89.7 & 93.5 & 89.2 \\
ConvNeXt-224-1K & 82.6 & 94.1 & 88.7 & 93.3 & 80.2 & 88.7 & 93.6 & 81.9 & 93.3 & 92.8 & 87.7 & 93.7 & 89.2 \\
ConvNeXt-224-22K & 81.9 & 94.8 & 89.0 & 93.6 & 80.4 & 87.3 & 94.0 & 82.7 & 93.0 & 92.9 & 88.2 & 93.6 & 89.3 \\
Swin-224-1K & 81.2 & 94.2 & 86.9 & 93.3 & 80.2 & 87.0 & 93.1 & 80.7 & 92.9 & 92.7 & 86.6 & 92.7 & 88.5 \\
Swin-224-22K & 81.1 & 94.2 & 87.5 & 93.9 & 77.7 & 87.2 & 92.9 & 81.3 & 93.4 & 92.1 & 85.9 & 92.1 & 88.3 \\
Swin-384-1K & 82.8 & 94.4 & 89.0 & 94.2 & 81.5 & 87.8 & 93.5 & 83.4 & 93.1 & 93.4 & 89.0 & 92.6 & 89.6 \\
Swin-384-22K & 82.2 & 94.1 & 89.5 & 93.7 & 79.1 & 87.7 & 94.1 & 82.2 & 93.2 & 93.3 & 88.0 & 92.8 & 89.2 \\
\hline
Max over members & 82.8 & 94.8 & 89.6 & 94.2 & 81.5 & 89.0 & 94.1 & 83.4 & 93.4 & 93.4 & 89.7 & 93.7 & 89.6 \\
Majority vote & 84.7 & 95.2 & 90.2 & 94.6 & 82.3 & 90.1 & 94.8 & 84.2 & 94.3 & 94.1 & 90.1 & 94.8 & 90.8 \\
Averaged & 84.8 & 95.2 & 90.4 & 94.6 & 82.5 & 90.0 & 94.8 & 84.6 & 94.3 & 94.2 & 90.5 & 94.7 & 90.9 \\
\bottomrule
\end{tabular}

\begin{tabular}{ l | c c c c c c c c c c c c | c }
\toprule
\multicolumn{14}{c}{Office-Home 1-shot No Augmentation} \\
\midrule
Backbone & R - C & R - P & R - A & P - R & P - C & P - A & A - P & A - C & A - R & C - R & C - A & C - P & Mean\\
\hline
ConvNeXt-384-1K & 82.4 & 95.1 & 90.5 & 94.2 & 79.2 & 89.6 & 94.6 & 82.0 & 93.5 & 92.8 & 90.1 & 94.1 & 89.8 \\
ConvNeXt-224-1K & 83.2 & 94.7 & 89.5 & 94.0 & 81.6 & 88.8 & 94.4 & 81.9 & 93.4 & 93.5 & 88.7 & 94.0 & 89.8 \\
ConvNeXt-224-22K & 82.7 & 94.7 & 89.2 & 94.2 & 81.5 & 87.9 & 94.0 & 82.1 & 94.0 & 93.3 & 88.9 & 94.0 & 89.7 \\
Swin-224-1K & 82.2 & 94.1 & 88.0 & 93.7 & 80.9 & 88.0 & 93.6 & 82.0 & 93.4 & 93.8 & 87.7 & 93.8 & 89.3 \\
Swin-224-22K & 83.4 & 94.4 & 87.6 & 93.6 & 79.2 & 87.5 & 94.7 & 81.2 & 93.5 & 93.1 & 87.6 & 92.4 & 89.0 \\
Swin-384-1K & 83.4 & 94.5 & 88.4 & 94.2 & 81.4 & 87.6 & 94.2 & 83.3 & 93.2 & 93.4 & 90.1 & 92.4 & 89.7 \\
Swin-384-22K & 83.3 & 94.1 & 89.2 & 94.0 & 80.1 & 88.7 & 94.6 & 83.2 & 93.2 & 93.1 & 89.3 & 93.3 & 89.7 \\
\hline
Max over members & 83.4 & 95.1 & 90.5 & 94.2 & 81.6 & 89.6 & 94.7 & 83.3 & 94.0 & 93.8 & 90.1 & 94.1 & 89.8 \\
Majority vote & 85.6 & 95.2 & 89.9 & 94.5 & 82.4 & 90.0 & 95.2 & 84.4 & 94.3 & 94.1 & 90.7 & 94.8 & 90.9 \\
Averaged & 85.7 & 95.2 & 90.0 & 94.5 & 82.6 & 90.2 & 95.1 & 84.6 & 94.3 & 94.1 & 90.6 & 94.9 & 91.0 \\
\bottomrule
\end{tabular}

\begin{tabular}{ l | c c c c c c c c c c c c | c }
\toprule
\multicolumn{14}{c}{Office-Home 1-shot Full Ensemble} \\
\midrule
Backbone & R - C & R - P & R - A & P - R & P - C & P - A & A - P & A - C & A - R & C - R & C - A & C - P & Mean\\
\hline
Majority vote  {\white jjjjjj } & 86.1 & 95.2 & 90.4 & 94.7 & 83.0 & 90.2 & 95.3 & 85.1 & 94.3 & 94.3 & 90.9 & 94.7 & 91.2 \\
Averaged & 86.2 & 95.2 & 90.4 & 94.7 & 82.9 & 90.4 & 95.2 & 85.4 & 94.4 & 94.3 & 91.1 & 94.7 & 91.2 \\
\bottomrule
\end{tabular}
}
\label{tab:office-home-1-shot-detailed}
\end{table}
\setlength\tabcolsep{6 pt}

\begin{table}
\caption{DomainNet 1-shot detailed results}
\centering
\setlength\tabcolsep{4.5 pt}
{ \small
\begin{tabular}{ l | c c c c c c c | c }
\toprule

\multicolumn{9}{c}{DomainNet 1-shot Grayscale} \\
\midrule
Backbone & R $\rightarrow$ C & R $\rightarrow$ P & P $\rightarrow$ C & C $\rightarrow$ S & S $\rightarrow$ P & R $\rightarrow$ S & P $\rightarrow$ R & Mean\\
\hline
ConvNeXt-384-1K & 77.0 & 82.3 & 76.4 & 74.5 & 82.0 & 74.1 & 89.7 & 79.4 \\
ConvNeXt-224-1K & 76.3 & 80.6 & 76.6 & 73.7 & 80.5 & 73.3 & 88.5 & 78.5 \\
ConvNeXt-224-22K & 76.8 & 80.5 & 76.1 & 71.7 & 80.3 & 70.9 & 89.0 & 77.9 \\
Swin-224-1K & 75.7 & 78.9 & 74.3 & 71.3 & 78.8 & 70.3 & 88.1 & 76.8 \\
Swin-224-22K & 76.6 & 79.0 & 75.5 & 70.2 & 79.1 & 69.4 & 88.5 & 76.9 \\
Swin-384-1K & 76.7 & 80.7 & 76.7 & 72.5 & 81.3 & 72.2 & 89.1 & 78.4 \\
Swin-384-22K & 77.3 & 81.1 & 77.7 & 72.0 & 81.4 & 70.8 & 89.4 & 78.5 \\
\hline
Max over members & 77.3 & 82.3 & 77.7 & 74.5 & 82.0 & 74.1 & 89.7 & 79.4 \\
Majority vote & 79.0 & 82.3 & 78.3 & 75.0 & 82.6 & 74.1 & 89.5 & 80.1 \\
Averaged & 79.2 & 82.4 & 78.5 & 75.0 & 82.5 & 74.3 & 89.6 & 80.2 \\
\bottomrule
\end{tabular}

\begin{tabular}{ l | c c c c c c c | c }
\toprule

\multicolumn{9}{c}{DomainNet 1-shot Perspective} \\
\midrule
Backbone & R $\rightarrow$ C & R $\rightarrow$ P & P $\rightarrow$ C & C $\rightarrow$ S & S $\rightarrow$ P & R $\rightarrow$ S & P $\rightarrow$ R & Mean\\
\hline
ConvNeXt-384-1K & 79.2 & 83.6 & 79.9 & 73.9 & 84.1 & 73.0 & 91.3 & 80.7 \\
ConvNeXt-224-1K & 78.3 & 81.8 & 78.9 & 72.4 & 81.3 & 70.9 & 90.8 & 79.2 \\
ConvNeXt-224-22K & 79.2 & 81.7 & 79.3 & 70.3 & 81.4 & 70.0 & 90.7 & 78.9 \\
Swin-224-1K & 78.9 & 81.0 & 79.8 & 69.6 & 81.3 & 69.2 & 90.7 & 78.7 \\
Swin-224-22K & 79.7 & 81.4 & 79.7 & 68.8 & 80.9 & 67.1 & 90.7 & 78.3 \\
Swin-384-1K & 80.8 & 82.1 & 80.8 & 70.2 & 82.4 & 70.6 & 91.3 & 79.7 \\
Swin-384-22K & 80.9 & 82.7 & 80.8 & 69.5 & 82.7 & 69.6 & 91.2 & 79.6 \\
\hline
Max over members & 80.9 & 83.6 & 80.8 & 73.9 & 84.1 & 73.0 & 91.3 & 80.7 \\
Majority vote & 81.3 & 83.7 & 82.4 & 73.2 & 84.1 & 72.4 & 91.6 & 81.2 \\
Averaged & 81.6 & 83.8 & 82.5 & 73.3 & 84.1 & 72.7 & 91.6 & 81.4 \\
\bottomrule
\end{tabular}

\begin{tabular}{ l | c c c c c c c | c }
\toprule

\multicolumn{9}{c}{DomainNet 1-shot RandAugment} \\
\midrule
Backbone & R $\rightarrow$ C & R $\rightarrow$ P & P $\rightarrow$ C & C $\rightarrow$ S & S $\rightarrow$ P & R $\rightarrow$ S & P $\rightarrow$ R & Mean\\
\hline
ConvNeXt-384-1K & 78.8 & 83.1 & 79.7 & 73.0 & 83.5 & 71.9 & 91.1 & 80.2 \\
ConvNeXt-224-1K & 78.7 & 81.8 & 78.5 & 72.1 & 81.6 & 71.2 & 90.8 & 79.2 \\
ConvNeXt-224-22K & 79.3 & 81.6 & 79.0 & 69.6 & 81.7 & 68.7 & 90.7 & 78.7 \\
Swin-224-1K & 78.5 & 80.8 & 78.9 & 69.0 & 80.9 & 68.1 & 90.6 & 78.2 \\
Swin-224-22K & 79.5 & 81.7 & 78.6 & 68.0 & 80.5 & 66.9 & 90.5 & 78.0 \\
Swin-384-1K & 80.2 & 82.3 & 79.1 & 70.4 & 82.4 & 70.5 & 91.1 & 79.4 \\
Swin-384-22K & 80.6 & 83.1 & 79.9 & 69.5 & 82.6 & 69.0 & 91.1 & 79.4 \\
\hline
Max over members & 80.6 & 83.1 & 79.9 & 73.0 & 83.5 & 71.9 & 91.1 & 80.2 \\
Majority vote & 81.9 & 84.0 & 81.8 & 73.7 & 84.4 & 72.5 & 91.7 & 81.4 \\
Averaged & 82.0 & 84.3 & 82.1 & 73.7 & 84.3 & 72.8 & 91.7 & 81.5 \\
\bottomrule
\end{tabular}

\begin{tabular}{ l | c c c c c c c | c }
\toprule

\multicolumn{9}{c}{DomainNet 1-shot No Augmentation} \\
\midrule
Backbone & R $\rightarrow$ C & R $\rightarrow$ P & P $\rightarrow$ C & C $\rightarrow$ S & S $\rightarrow$ P & R $\rightarrow$ S & P $\rightarrow$ R & Mean\\
\hline
ConvNeXt-384-1K & 80.1 & 83.6 & 80.0 & 72.9 & 84.1 & 72.2 & 91.3 & 80.6 \\
ConvNeXt-224-1K & 80.1 & 82.6 & 78.7 & 72.5 & 82.6 & 71.7 & 91.1 & 79.9 \\
ConvNeXt-224-22K & 79.7 & 82.5 & 79.2 & 70.5 & 81.5 & 69.6 & 91.0 & 79.1 \\
Swin-224-1K & 79.7 & 81.4 & 79.7 & 69.9 & 81.8 & 69.2 & 90.9 & 79.0 \\
Swin-224-22K & 80.5 & 81.9 & 79.7 & 68.6 & 81.1 & 67.7 & 90.8 & 78.6 \\
Swin-384-1K & 80.8 & 82.7 & 80.6 & 70.8 & 83.4 & 70.8 & 91.3 & 80.0 \\
Swin-384-22K & 81.4 & 83.3 & 80.8 & 69.9 & 83.2 & 69.5 & 91.3 & 79.9 \\
\hline
Max over members & 81.4 & 83.6 & 80.8 & 72.9 & 84.1 & 72.2 & 91.3 & 80.6 \\
Majority vote & 82.2 & 84.0 & 82.0 & 73.5 & 84.5 & 72.7 & 91.7 & 81.5 \\
Averaged & 82.3 & 84.2 & 82.2 & 73.6 & 84.6 & 72.8 & 91.7 & 81.6 \\
\bottomrule
\end{tabular}

\begin{tabular}{ l | c c c c c c c | c }
\toprule

\multicolumn{9}{c}{DomainNet 1-shot Full Ensemble} \\
\midrule
Backbone & R $\rightarrow$ C & R $\rightarrow$ P & P $\rightarrow$ C & C $\rightarrow$ S & S $\rightarrow$ P & R $\rightarrow$ S & P $\rightarrow$ R & Mean\\
\hline
Majority vote {\white jjjjjj }  & 82.2 & 84.4 & 82.5 & 74.6 & 84.9 & 74.0 & 91.7 & 82.0 \\
Averaged & 82.4 & 84.5 & 82.6 & 74.6 & 84.8 & 74.0 & 91.7 & 82.1 \\
\bottomrule
\end{tabular}
}
\label{tab:domainnet-1-shot-detailed}
\end{table}
\setlength\tabcolsep{6 pt}

\begin{table}
\caption{Office-Home zero-shot (UDA) detailed results}
\centering
\setlength\tabcolsep{2.5 pt}
{ \small
\begin{tabular}{ l | c c c c c c c c c c c c | c }
\toprule
\multicolumn{14}{c}{Office-Home zero-shot (UDA) Grayscale} \\
\midrule
Backbone & R - C & R - P & R - A & P - R & P - C & P - A & A - P & A - C & A - R & C - R & C - A & C - P & Mean\\
\hline
ConvNeXt-384-1K & 79.2 & 94.5 & 89.1 & 93.8 & 77.9 & 87.8 & 93.0 & 77.9 & 92.9 & 92.0 & 88.3 & 92.1 & 88.2 \\
ConvNeXt-224-1K & 79.2 & 92.8 & 87.8 & 93.5 & 78.3 & 86.1 & 90.9 & 77.7 & 92.8 & 92.7 & 87.4 & 92.2 & 87.6 \\
ConvNeXt-224-22K & 78.8 & 93.9 & 88.8 & 93.5 & 76.6 & 85.3 & 90.8 & 78.0 & 92.7 & 92.2 & 87.3 & 91.2 & 87.4 \\
Swin-224-1K & 77.9 & 92.5 & 86.3 & 92.7 & 76.9 & 84.1 & 89.7 & 77.6 & 92.0 & 92.1 & 85.1 & 91.1 & 86.5 \\
Swin-224-22K & 78.7 & 93.2 & 86.3 & 92.7 & 75.4 & 83.8 & 90.1 & 78.2 & 92.6 & 92.0 & 85.4 & 90.0 & 86.5 \\
Swin-384-1K & 80.3 & 93.3 & 86.9 & 93.6 & 78.6 & 84.6 & 91.1 & 78.7 & 92.2 & 92.6 & 86.7 & 90.9 & 87.5 \\
Swin-384-22K & 79.6 & 92.9 & 88.0 & 92.7 & 76.4 & 84.8 & 91.6 & 79.5 & 92.4 & 92.6 & 86.4 & 90.2 & 87.3 \\
\hline
Max over members & 80.3 & 94.5 & 89.1 & 93.8 & 78.6 & 87.8 & 93.0 & 79.5 & 92.9 & 92.7 & 88.3 & 92.2 & 88.2 \\
Majority vote & 81.4 & 94.2 & 88.6 & 93.9 & 79.5 & 86.5 & 92.0 & 80.5 & 93.4 & 93.5 & 88.3 & 92.5 & 88.7 \\
Averaged & 81.5 & 94.2 & 88.5 & 93.9 & 79.7 & 86.7 & 92.4 & 80.7 & 93.4 & 93.6 & 88.7 & 92.5 & 88.8 \\
\bottomrule
\end{tabular}

\begin{tabular}{ l | c c c c c c c c c c c c | c }
\toprule
\multicolumn{14}{c}{Office-Home zero-shot (UDA) Perspective} \\
\midrule
Backbone & R - C & R - P & R - A & P - R & P - C & P - A & A - P & A - C & A - R & C - R & C - A & C - P & Mean\\
\hline
ConvNeXt-384-1K & 83.6 & 95.2 & 89.6 & 94.5 & 81.5 & 88.9 & 92.7 & 83.1 & 93.5 & 93.5 & 89.8 & 94.3 & 90.0 \\
ConvNeXt-224-1K & 83.8 & 94.4 & 88.7 & 93.6 & 82.0 & 87.6 & 93.5 & 83.6 & 93.0 & 93.3 & 89.1 & 92.8 & 89.6 \\
ConvNeXt-224-22K & 84.1 & 94.3 & 88.7 & 93.8 & 80.2 & 87.6 & 92.5 & 83.6 & 93.3 & 93.4 & 88.7 & 92.6 & 89.4 \\
Swin-224-1K & 84.0 & 94.7 & 87.8 & 93.9 & 81.6 & 86.0 & 91.9 & 83.6 & 93.4 & 93.2 & 87.6 & 90.0 & 89.0 \\
Swin-224-22K & 83.0 & 94.5 & 88.0 & 94.0 & 81.0 & 86.3 & 92.4 & 82.5 & 93.5 & 92.4 & 87.3 & 91.6 & 88.9 \\
Swin-384-1K & 84.5 & 94.0 & 88.2 & 94.2 & 82.2 & 87.7 & 92.7 & 84.9 & 93.8 & 93.8 & 88.8 & 92.8 & 89.8 \\
Swin-384-22K & 83.6 & 93.8 & 89.2 & 94.1 & 81.1 & 87.1 & 92.2 & 83.9 & 93.8 & 93.5 & 88.8 & 92.1 & 89.4 \\
\hline
Max over members & 84.5 & 95.2 & 89.6 & 94.5 & 82.2 & 88.9 & 93.5 & 84.9 & 93.8 & 93.8 & 89.8 & 94.3 & 90.0 \\
Majority vote & 85.7 & 95.2 & 89.9 & 94.6 & 83.9 & 88.9 & 93.6 & 85.5 & 94.2 & 94.3 & 90.5 & 93.3 & 90.8 \\
Averaged & 85.9 & 95.3 & 89.9 & 94.8 & 83.8 & 88.9 & 93.8 & 85.7 & 94.3 & 94.4 & 91.1 & 93.2 & 90.9 \\
\bottomrule
\end{tabular}

\begin{tabular}{ l | c c c c c c c c c c c c | c }
\toprule
\multicolumn{14}{c}{Office-Home zero-shot (UDA) RandAugment} \\
\midrule
Backbone & R - C & R - P & R - A & P - R & P - C & P - A & A - P & A - C & A - R & C - R & C - A & C - P & Mean\\
\hline
ConvNeXt-384-1K & 80.6 & 94.8 & 89.0 & 94.4 & 77.9 & 87.4 & 93.9 & 80.0 & 94.0 & 92.9 & 88.5 & 92.5 & 88.8 \\
ConvNeXt-224-1K & 81.5 & 93.8 & 88.7 & 93.4 & 79.8 & 86.6 & 92.6 & 80.4 & 92.9 & 92.5 & 88.1 & 93.4 & 88.6 \\
ConvNeXt-224-22K & 82.1 & 94.3 & 88.4 & 93.6 & 79.6 & 87.1 & 91.9 & 79.3 & 93.0 & 92.8 & 87.2 & 93.6 & 88.6 \\
Swin-224-1K & 79.7 & 93.4 & 86.9 & 92.7 & 78.1 & 86.2 & 91.2 & 79.6 & 93.1 & 92.7 & 86.3 & 90.4 & 87.5 \\
Swin-224-22K & 80.2 & 93.7 & 86.9 & 93.3 & 78.3 & 85.9 & 92.6 & 80.3 & 93.6 & 92.4 & 86.7 & 90.7 & 87.9 \\
Swin-384-1K & 81.1 & 94.0 & 88.0 & 93.9 & 80.2 & 86.5 & 92.4 & 80.8 & 93.1 & 92.9 & 89.0 & 91.8 & 88.6 \\
Swin-384-22K & 82.3 & 93.2 & 89.0 & 93.4 & 79.6 & 86.5 & 93.2 & 80.8 & 93.7 & 93.0 & 87.4 & 91.9 & 88.7 \\
\hline
Max over members & 82.3 & 94.8 & 89.0 & 94.4 & 80.2 & 87.4 & 93.9 & 80.8 & 94.0 & 93.0 & 89.0 & 93.6 & 88.8 \\
Majority vote & 84.0 & 94.7 & 89.8 & 94.3 & 82.4 & 89.3 & 94.1 & 82.7 & 94.2 & 94.2 & 90.2 & 93.8 & 90.3 \\
Averaged & 84.3 & 94.9 & 89.8 & 94.5 & 82.4 & 89.1 & 94.0 & 82.9 & 94.2 & 94.1 & 90.4 & 93.9 & 90.4 \\
\bottomrule
\end{tabular}

\begin{tabular}{ l | c c c c c c c c c c c c | c }
\toprule
\multicolumn{14}{c}{Office-Home zero-shot (UDA) No Augmentation} \\
\midrule
Backbone & R - C & R - P & R - A & P - R & P - C & P - A & A - P & A - C & A - R & C - R & C - A & C - P & Mean\\
\hline
ConvNeXt-384-1K & 81.3 & 94.6 & 90.2 & 94.3 & 79.4 & 88.2 & 92.9 & 80.9 & 93.4 & 92.7 & 89.3 & 93.3 & 89.2 \\
ConvNeXt-224-1K & 82.4 & 94.8 & 89.1 & 93.8 & 81.0 & 87.6 & 93.5 & 79.9 & 93.6 & 93.3 & 88.7 & 93.4 & 89.3 \\
ConvNeXt-224-22K & 81.6 & 94.9 & 89.3 & 93.9 & 80.1 & 87.5 & 91.8 & 79.7 & 93.6 & 92.7 & 88.6 & 93.8 & 89.0 \\
Swin-224-1K & 81.4 & 93.9 & 87.4 & 93.8 & 80.0 & 86.5 & 91.9 & 80.8 & 93.6 & 93.2 & 87.1 & 92.0 & 88.5 \\
Swin-224-22K & 82.7 & 94.3 & 88.3 & 93.8 & 78.5 & 85.9 & 91.9 & 80.5 & 93.7 & 93.2 & 87.7 & 91.9 & 88.5 \\
Swin-384-1K & 82.7 & 94.4 & 87.5 & 94.2 & 81.1 & 88.0 & 92.8 & 81.4 & 93.4 & 93.3 & 89.5 & 92.8 & 89.3 \\
Swin-384-22K & 82.6 & 93.9 & 89.2 & 94.0 & 78.7 & 87.5 & 93.8 & 81.5 & 93.7 & 93.5 & 89.3 & 91.9 & 89.1 \\
\hline
Max over members & 82.7 & 94.9 & 90.2 & 94.3 & 81.1 & 88.2 & 93.8 & 81.5 & 93.7 & 93.5 & 89.5 & 93.8 & 89.3 \\
Majority vote & 85.0 & 95.1 & 89.9 & 94.5 & 81.9 & 89.2 & 94.1 & 82.3 & 94.2 & 94.3 & 90.4 & 93.7 & 90.4 \\
Averaged & 85.0 & 95.1 & 89.8 & 94.5 & 82.0 & 89.3 & 94.2 & 82.5 & 94.2 & 94.2 & 90.3 & 93.7 & 90.4 \\
\bottomrule
\end{tabular}

\begin{tabular}{ l | c c c c c c c c c c c c | c }
\toprule
\multicolumn{14}{c}{Office-Home zero-shot (UDA) Full Ensemble} \\
\midrule
Backbone & R - C & R - P & R - A & P - R & P - C & P - A & A - P & A - C & A - R & C - R & C - A & C - P & Mean\\
\hline
Majority vote  {\white jjjjjj } & 85.7 & 95.2 & 90.3 & 94.6 & 83.0 & 89.5 & 93.9 & 83.5 & 94.2 & 94.4 & 90.6 & 93.4 & 90.7 \\
Averaged & 85.7 & 95.3 & 90.3 & 94.6 & 83.0 & 89.6 & 94.0 & 84.0 & 94.3 & 94.4 & 90.7 & 93.6 & 90.8 \\

\bottomrule
\end{tabular}
}
\label{tab:office-home-0-shot-detailed}
\end{table}
\setlength\tabcolsep{6 pt}

\end{document}